\documentclass[journal,comsoc]{IEEEtran}

\usepackage{CJKutf8}

\usepackage{framed}
\usepackage{color}
\definecolor{shadecolor}{rgb}{0.72,0.72,0.72}
\definecolor{ashgrey}{rgb}{0.7, 0.75, 0.71}
\definecolor{battleshipgrey}{rgb}{0.52, 0.52, 0.51}	
\definecolor{cadetgrey}{rgb}{0.57, 0.64, 0.69}	
\definecolor{coolgrey}{rgb}{0.55, 0.57, 0.67}

\usepackage{xcolor}
\usepackage[normalem]{ulem} 
\newcommand\hl{\bgroup\markoverwith
	{\textcolor{shadecolor}{\rule[-.65ex]{2pt}{4.7ex}}}\ULon}

\usepackage{tipa}
\usepackage{threeparttable}
\usepackage{booktabs}
\usepackage{multirow}

\usepackage[T1]{fontenc}
\usepackage{graphicx}
\usepackage{bm}
\usepackage[noend]{algpseudocode}
\usepackage{algorithmicx,algorithm}
\usepackage{threeparttable}
\usepackage[nocompress,sort]{cite}

\usepackage{stackengine}

\ifCLASSINFOpdf
\else
\fi
\usepackage{amsmath}
\usepackage[cmintegrals]{newtxmath}
\ifCLASSOPTIONcompsoc
  \usepackage[caption=false,font=normalsize,labelfont=sf,textfont=sf]{subfig}
\else
  \usepackage[caption=false,font=footnotesize]{subfig}
\fi
\usepackage{hyperref}
\hypersetup{hidelinks}

\def\equationautorefname#1#2\null{
	Eq. (#2\null)
}

\begin{document}
%
\title{Fast End-to-End Speech Recognition via Non-Autoregressive Models and Cross-Modal Knowledge Transferring from BERT}
%
%
%

\author{Ye~Bai,~\IEEEmembership{Student Member,~IEEE,}
	Jiangyan~Yi,~\IEEEmembership{Member,~IEEE,}
	Jianhua~Tao,~\IEEEmembership{Senior Member,~IEEE}
	Zhengkun~Tian,~\IEEEmembership{Student Member,~IEEE}
	Zhengqi~Wen,~\IEEEmembership{Member,~IEEE}
	Shuai~Zhang,~\IEEEmembership{Student Member,~IEEE}
	\thanks{Ye Bai, Zhengkun Tian, Jianhua Tao and Shuai Zhang are with the University of Chinese Academy of Sciences, Beijing 100190, China, and with NLPR, Institute of Automation, Chinese Academy of Sciences, Beijing 100190, China. (e-mail: baiye2016@ia.ac.cn;	zhengkun.tian@nlpr.ia.ac.cn; shuai.zhang@nlpr.ia.ac.cn).}
	\thanks{Jiangyan Yi and Zhengqi Wen are with NLPR, Institute of Automation, Chinese Academy of Sciences, Beijing 100190, China (e-mail: jiangyan.yi@nlpr.ia.ac.cn; zqwen@nlpr.ia.ac.cn).}
	\thanks{Jianhua Tao is also with NLPR, Institute of Automation, Chinese Academy of Sciences, Beijing 100190, China, and the CAS Center for Excellence
		in Brain Science and Intelligence Technology, Beijing 100190, China (e-mail: jhtao@nlpr.ia.ac.cn). (Corresponding authors: Jiangyan Yi; Jianhua Tao.)}

}

%
%

\markboth{Journal of \LaTeX\ Class Files,~Vol.~14, No.~8, August~2015}%
{Shell \MakeLowercase{\textit{et al.}}: Bare Demo of IEEEtran.cls for IEEE Communications Society Journals}
%



\maketitle
\begin{abstract}
Attention-based encoder-decoder (AED) models have achieved promising performance in speech recognition. However, because the decoder predicts text tokens (such as characters or words) in an autoregressive manner, it is difficult for an AED model to predict all tokens in parallel. This makes the inference speed relatively slow. In contrast, we propose an end-to-end non-autoregressive speech recognition model called LASO (Listen Attentively, and Spell Once). The model aggregates encoded speech features into the hidden representations corresponding to each token with attention mechanisms. Thus, the model can capture the token relations by self-attention on the aggregated hidden representations from the whole speech signal rather than autoregressive modeling on tokens. Without explicitly autoregressive language modeling, this model predicts all tokens in the sequence in parallel so that the inference is efficient. Moreover, we propose a cross-modal transfer learning method to use a text-modal language model to improve the performance of speech-modal LASO by aligning token semantics. We conduct experiments on two scales of public Chinese speech datasets AISHELL-1 and AISHELL-2. Experimental results show that our proposed model achieves a speedup of about $50\times$ and competitive performance, compared with the autoregressive transformer models. And the cross-modal knowledge transferring from the text-modal model can improve the performance of the speech-modal model. 
\end{abstract}

\begin{IEEEkeywords}
speech recognition, fast, end-to-end, non-autoregressive, attention, BERT, cross-modal, transfer learning
\end{IEEEkeywords}

%
\IEEEpeerreviewmaketitle

%
%
%
%


\section{Introduction}
\IEEEPARstart{D}{eep} learning has significantly improved the performance of automatic speech recognition (ASR). Conventionally, an ASR system consists of an acoustic model (AM), a pronunciation lexicon, and a language model (LM). The deep neural network (DNN) is used to model observation probabilities of the hidden Markov models (HMMs) \cite{hinton2012deep,yu2016automatic}. This DNN-HMM hybrid approach has achieved success in ASR. However, the pipeline of a DNN-HMM hybrid system usually requires training Gaussian mixture model based HMMs (GMM-HMM) for generating frame-level alignments and tying states. Building the pronunciation lexicon requires the knowledge of experts in phonetics. This complexity of the building pipeline limits the development of an ASR system. Moreover, the different building procedures of the AM and the LM make the system difficult to be optimized jointly. Thus, the possible error accumulation in the pipeline influences the performance of a hybrid ASR system.

Pure neural network based end-to-end (E2E) ASR systems attract interests of researchers these years \cite{chorowski2015attention,bahdanau2016endtoend,chan2016listen,kim2017joint,graves2012sequence}. Different from the hybrid ASR systems, these systems use one DNN to model acoustic and language simultaneously so that the network can be optimized with back-propagation algorithms in an E2E manner. In particular, attention-based encoder-decoder (AED) models have achieved promising performance in ASR \cite{chiu2018state,luscher2019rwth}. The AED models first encode the acoustic feature sequence into latent representations with an encoder. With the latent representations, the decoder predicts the text token sequence step-by-step. The attention mechanism queries a proper latent vector from the outputs of the encoder for the decoder to predict. However, even with the non-recurrent structure which can be implemented in parallel \cite{dong2018speech,zhou2018syllable}, the recognition speed limits the deployment of AED models in real-world applications. Two main reasons influence inference speed: 
\begin{itemize}
	\item First, the encoder encodes the whole utterance, so the decoder starts inference after the user speaks out the whole utterance;
	\item Second, multi-pass forward propagation of the decoder costs much time during beam-search.
\end{itemize}

\begin{figure}[!t]\centering
	\includegraphics[width=0.9\columnwidth]{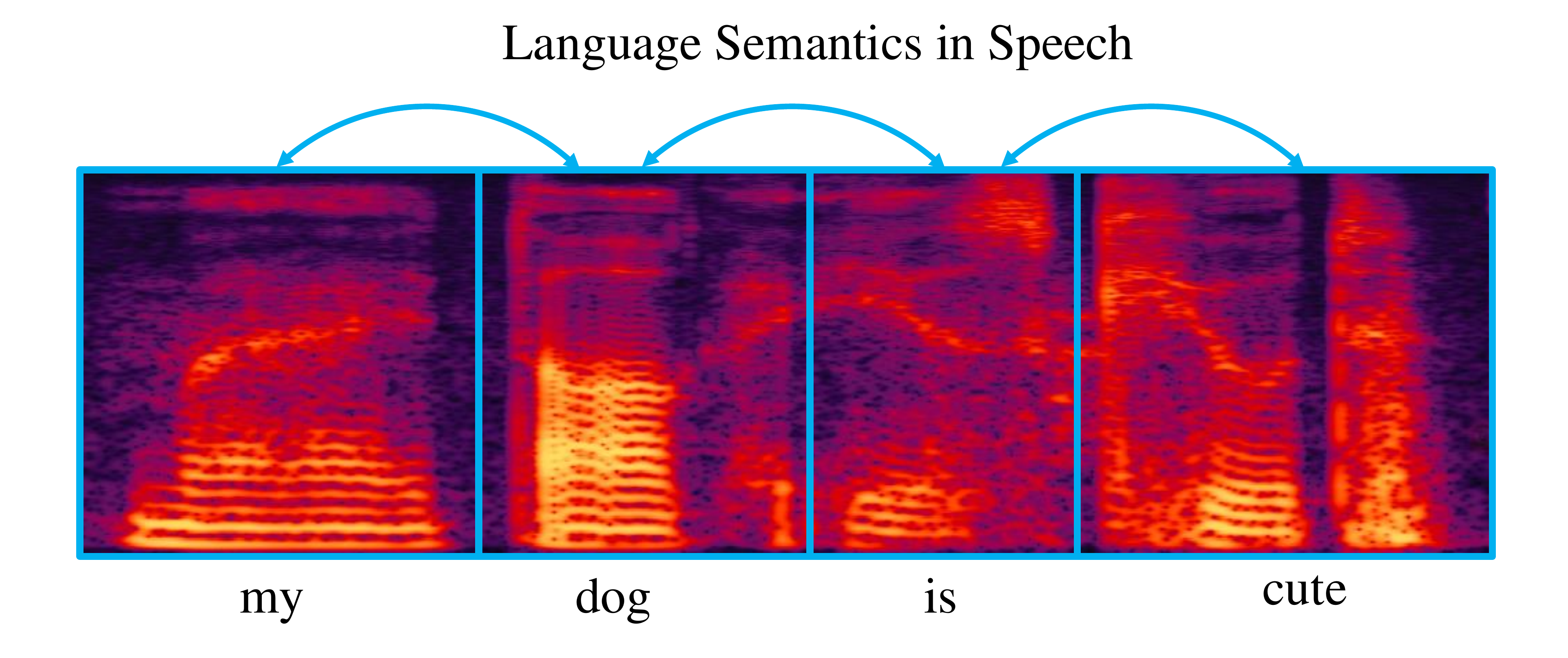} 
	\caption{A spectrogram of an example utterance. A word corresponds to a segment in the speech signal. The relationships among the segments can be seen as the relationships among the corresponding tokens, which are referred to as language semantics in this paper.} 
	\label{fig:eg}
	\vspace{-10pt}
\end{figure}

Several work focus on the first problem to make the system can generate the token sequence in a streaming manner. Monotonic attention mechanism \cite{raffel2017online,mocha2018} enforces the attention alignments to be monotonic. Therefore, the encoder only encodes a local chunk in the acoustic feature sequence, and the decoder predicts the next token without the future context in the acoustic feature sequence. Triggered-attention systems \cite{moritz2019triggered,moritz2019streaming} use connectionist temporal classification (CTC) spikes to segment the acoustic feature sequence adaptively, and the decoder predicts the next token when it is activated by a spike. Transducer-based models \cite{graves2012sequence,tian2019self,tian2020synchronous} use an extra blank token and marginalize all possible alignments, so the model can immediately predict the next token when the encoder accumulates enough information. All these models can generate a token sequence in a steaming manner and show promising results. However, they limit the models using local information in the speech sequence. It potentially ignores global semantic relationships in the speech sequence. The global semantic relationships contain not only the relationships among acoustic frames but also the relationships among tokens \cite{Chung2018}, as shown in \autoref{fig:eg}.

In this paper, we aim to address the second problem, i.e., we would like to generate the token sequence without beam-search. We propose an attention-based feedforward neural network model for non-autoregressive speech recognition called ``LASO'' (Listen Attentively, and Spell Once). The LASO model first uses an encoder to encode the whole acoustic sequence into high-level representations. Then, the proposed \textit{position dependent summarizer} (PDS) module queries the latent representation corresponding to the token position from the outputs of the encoder. It bridges the length gap between the speech and the token sequence. At last, the decoder further refines the representations and predicts a token for each position. Because the prediction of one token does not depend on another token, beam-search is not used. And because the network is a non-recurrent feedforward structure, it can be implemented in parallel. To further improve the ability to capture the semantic relationship of LASO (especially for the decoder), we propose to use a cross-modal knowledge transferring method. Specifically, we align the semantic spaces of the hidden representations of the LASO and the pre-trained large-scale language model BERT \cite{devlin2018bert}. We use the teacher-student learning based knowledge transferring method \cite{bai2019learn} to leverage knowledge in BERT. We conduct experiments on two publicly available Chinese Mandarin datasets to evaluate the proposed methods with different data sizes. The experiments demonstrate that our proposed method achieves a competitive performance and efficiency.

The contributions of this work are summarized as follows.
\begin{enumerate}
	\item We propose a non-autoregressive attention-based feedforward neural network LASO for speech recognition. A PDS module is proposed to convert a variant-length speech sequence to a fixed-length sequence. Each representation in the fixed-length sequence corresponds to the position of a token so that the ASR is considered as a position-wise classification. A decoder, which plays a role of bidirectional LM, further captures the relationship among these representations with self-attention. LASO leverages the whole context of the speech and generates each token in the sequence in parallel. The experiments demonstrate that our proposed model achieves high efficiency and competitive performance. 
	\item We propose a cross-modal knowledge transferring method from BERT for improving the performance of LASO. The experiments demonstrate the effectiveness of the knowledge transferring from BERT. The results also show that the speech signals have similar internal structures with corresponding text so that knowledge from BERT can benefit the non-autoregressive ASR model.
	\item We present detailed visualization of the model. The visualization results show that the proposed PDS module can attend to specific encoded acoustic representations and generates meaningful hidden representations corresponding to tokens from speech. And the decoder can capture token relationships from the aggregated hidden representations. 
\end{enumerate}

Compared with the preliminary version \cite{bai2020listen}, the new content in this paper includes leveraging pre-trained BERT models to further improve the performance, graph-based view of the proposed model, more detailed experiments on large-scale datasets, detailed visualization and analysis for better understanding the models. The rest of the paper is organized as follows. \autoref{sec:bg} briefly compares the autoregressive AED models and the non-autoregressive AED models. \autoref{sec:cls} re-formulates the speech recognition as a position-wise classification problem. \autoref{sec:laso} describes the proposed LASO model. \autoref{sec:learning} describes how to train the model and how we distill the knowledge from BERT. \autoref{sec:inference} introduces how the model generates a sentence. \autoref{sec:rel} compares this work with previous related work. \autoref{sec:setup} and \autoref{sec:res} present setup and results of experiments, respectively. \autoref{sec:discuss} discusses this paper. At last, \autoref{sec:conc} concludes this paper and presents future work.

\begin{figure*}[!t]\centering
	\includegraphics[width=0.9\linewidth]{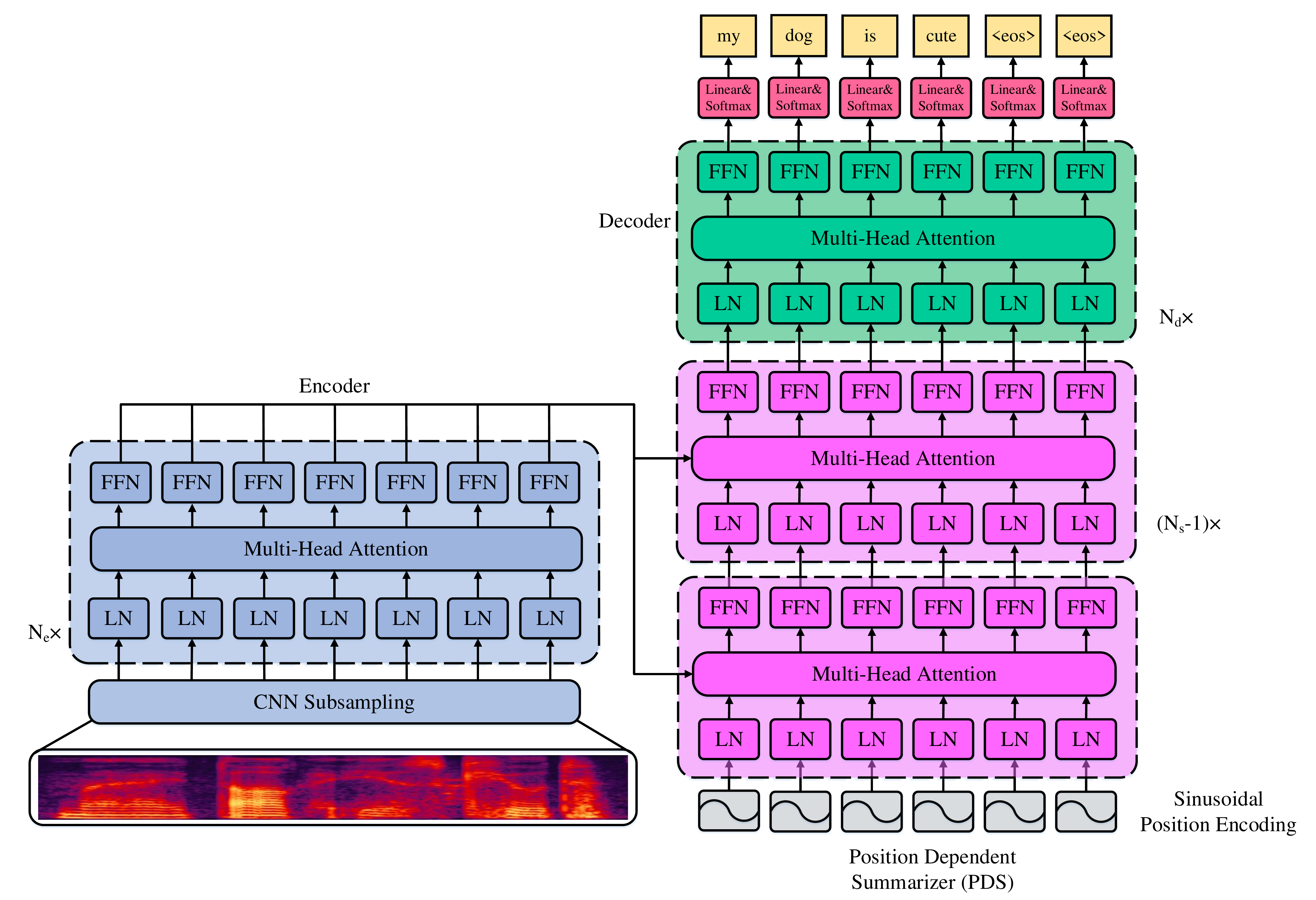} 
	\caption{An illustration of the proposed LASO model. The LASO consists of an encoder, a position-dependent summarizer (PDS), and a decoder. All three modules are composed of basic attention blocks, which consist of multi-head attention and a position-wise feedforward network (FFN in the figure). We first use a CNN to subsample the acoustic feature sequence. Then, the encoder extracts high-level representations from the subsampled sequence. The PDS queries the high-level acoustic representations corresponding to each token position. Then, the decoder further refines the language semantics. For each position, a probability distribution over the vocabulary is computed with a softmax function. And during inference, we select the most likely token at each position to form the token sequence. The extra tokens \texttt{<eos>} are removed in the token sequence. Each attention block includes layer normalization (LN in the figure) and a residual connection. And we add sinusoidal position encodings to the subsampled acoustic feature sequence. The whole architecture is non-recurrent, so it can be implemented in parallel for fast inference.} 
	\label{fig:arch}
	\vspace{-10pt}
\end{figure*}

\section{Background: Autoregressive Models vs. Non-Autoregressive Models}
\label{sec:bg}
In this section, we introduce the background of autoregressive AED models (ARM) and non-autoregressive AED models (NARM). We compare these two paradigms to better introduce the proposed method in the rest sections.

\vspace{-5pt}
\subsection{Autoregressive AED Models}
The ARM predicts the next token based on the previously generated tokens. That is, for a speech-text pair $(X, Y)$, the ARM factorizes the conditional probability $P(Y|X)$ with the chain rule:
\begin{equation}
\label{eq:arm}
P_{\text{ARM}}(Y|X) = P(y_1|X)\prod_{j=2}^{L} P(y_j|y_{<j}, X),
\end{equation}
where $X=[x_1, \cdots, x_T]$ is the acoustic feature sequence, each $x$ denotes a feature vector, $Y=[y_1, \cdots, y_L]$ is the text token sequence, and $y_{<j}=[y_1, \cdots, y_{j-1}]$ denotes the previous context of the token $y_j$. The token can be a word or a sub-word (phone, character, or word piece). This can be seen as a conditional language model, i.e., the model estimates the probability of the token sequence given the acoustic feature sequence. Typically, \autoref{eq:arm} is implemented with a neural network, which is an encoder-decoder architecture. The encoder encodes the acoustic feature sequence, and the decoder computes $P(y_j|y_{<j}, X)$ step-by-step. 

To find the token sequence which has the highest probability approximately, a beam-search algorithm is used during inference. The decoder maintains a beam of potential candidates of the token sequence which have high probabilities. Thus, the decoder has to forward propagate these candidates. In addition, the generation of each token depends on the previously generated tokens, so it is difficult to implement parallel generation.

\vspace{-5pt}
\subsection{Non-Autoregressive AED Models}
Different from the ARM, the NARM predicts each token without dependence on other tokens. Specifically, the NARM assumes the conditional independence on tokens:
\begin{equation}
\label{eq:narm}
P_{\text{NARM}}(Y|X) = \prod_{j=1}^{L} P(y_j|X).
\end{equation}
Because each probability does not depend on the other tokens, parallel implementation is possible. Another view of \autoref{eq:narm} is to process each token independently rather than to process the production. The details are described in \autoref{sec:cls}.

Conventionally, the relationships among tokens are considered an important factor for the token sequence generation. We refer to this relationship as \textit{language semantics} in this paper. The previous non-autoregressive model CTC assumes conditional independence on token sequence \cite{graves2006connectionist}. However, to achieve good performance, the CTC-based systems use n-gram LMs for modeling the language semantics \cite{miao2015eesen}. And the advanced version transducer models \cite{graves2012sequence} use a neural network to model the language semantics. An AED model captures the language semantics by the decoder \cite{chorowski2015attention,bahdanau2016endtoend,chan2016listen}. Different from them, in this paper, we propose to use a self-attention mechanism to model the implicit language semantics, which compensates for the loss of the explicit autoregressive language model. The details are described in \autoref{sec:laso}.

\section{ASR as Position-wise Classification}
\label{sec:cls}
In this paper, we give a new perspective on the speech recognition problem. The basic idea is that the language semantics is implicitly contained in the speech signal. \autoref{fig:eg} shows a spectrogram of an utterance ``my dog is cute''. Each segment corresponds to a word in the utterance. We observe that the language semantics, i.e., the relationships among the tokens, is expressed among the segments implicitly. Thus, if the speech signal of the whole utterance is available, we can leverage this implicit language semantic to improve the performance of ASR. 

Based on the above observations, we consider the ASR problem as position-wise classification, given the whole speech utterance. Namely, we use the whole acoustic feature sequence, including explicit acoustic characteristics and implicit language semantics, to predict one token, but not estimate the probability of the token sequence. When all the tokens are predicted, we simply put them together as the recognition result. Formally, we predict the following probability:
\begin{equation}
\label{eq:pcp}
P(y_j|X) = f(X), \quad j=1,\cdots,L,
\end{equation}
where $X$ denotes the whole acoustic feature sequence, $y_j$ denotes a token in the token sequence, $L$ is the length of the token sequence, and $f$ is some non-linear function. In this paper, we use the proposed feedforward neural network LASO as the non-linear function $f$. Usually, the length of the token sequence is unknown in advance. We use a simple way to tackle this. We set $L$ as a big enough number, and the tail of the token sequence will be predicted to the filler token.

\begin{figure}[!t]\centering
	\includegraphics[width=1.0\columnwidth]{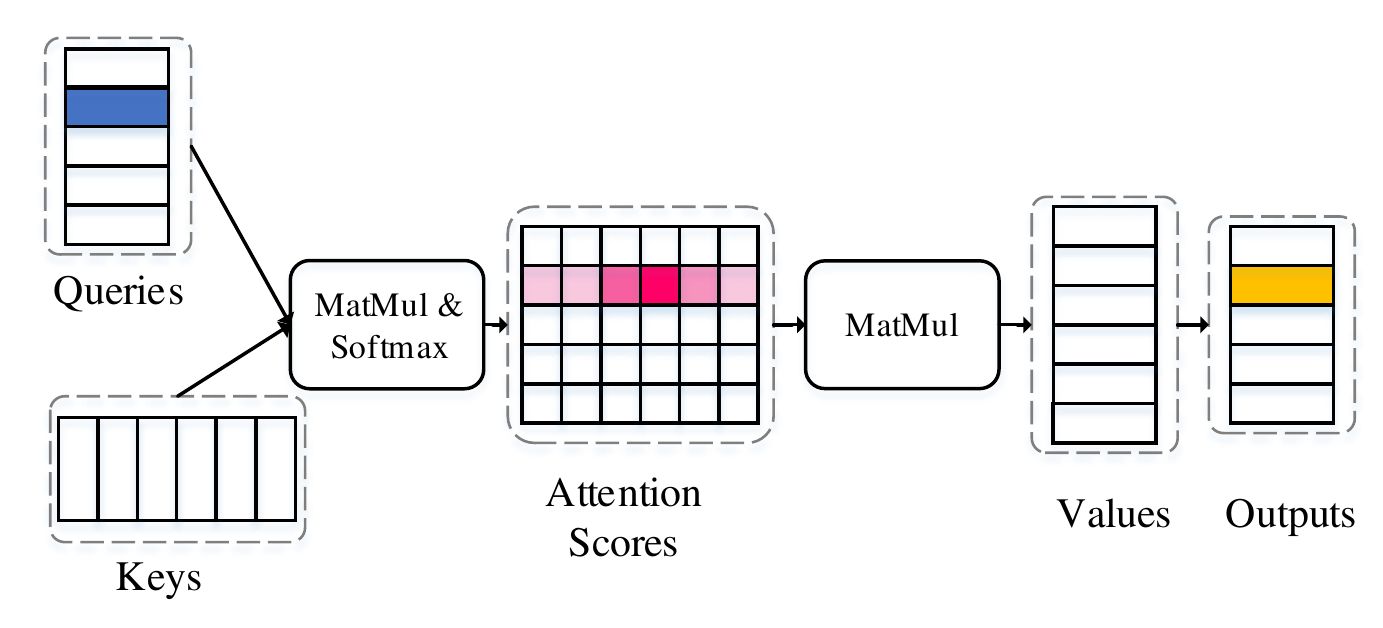} 
	\caption{An illustration of dot-product attention. The queries and the keys are used to compute the attention scores with matrix multiplication and the softmax function. The attention scores are used as weights to fuse values.} 
	\label{fig:attention}
	\vspace{-10pt}
\end{figure}

\section{The Proposed LASO Model}
\label{sec:laso}

In this section, we introduce the proposed LASO model. The architecture is shown in \autoref{fig:arch}. The encoder encodes the acoustic feature sequence into high-level representations. The PDS queries the high-level representation corresponding to each token position. Another purpose of the PDS is to bridge the length gap between the speech and the token sequence. The decoder further captures language semantics from the outputs of the PDS. At last, the probability distributions over the vocabulary are computed with the linear transformations and the softmax functions. For inference, the most likely token at each position is selected. For the token sequence whose length is shorter than $L$, the tail is filled with the filler token \texttt{<eos>}. This also predicts the length of the token sequence automatically. These filler tokens are easily removed.

The whole network is a feedforward structure so that it can be implemented in parallel to make the inference fast. We introduce the details of the basic attention block and each module in the rest of this section. We also introduce a graph-based view of the whole model.

\begin{figure}[!t]\centering
	\includegraphics[width=0.5\columnwidth]{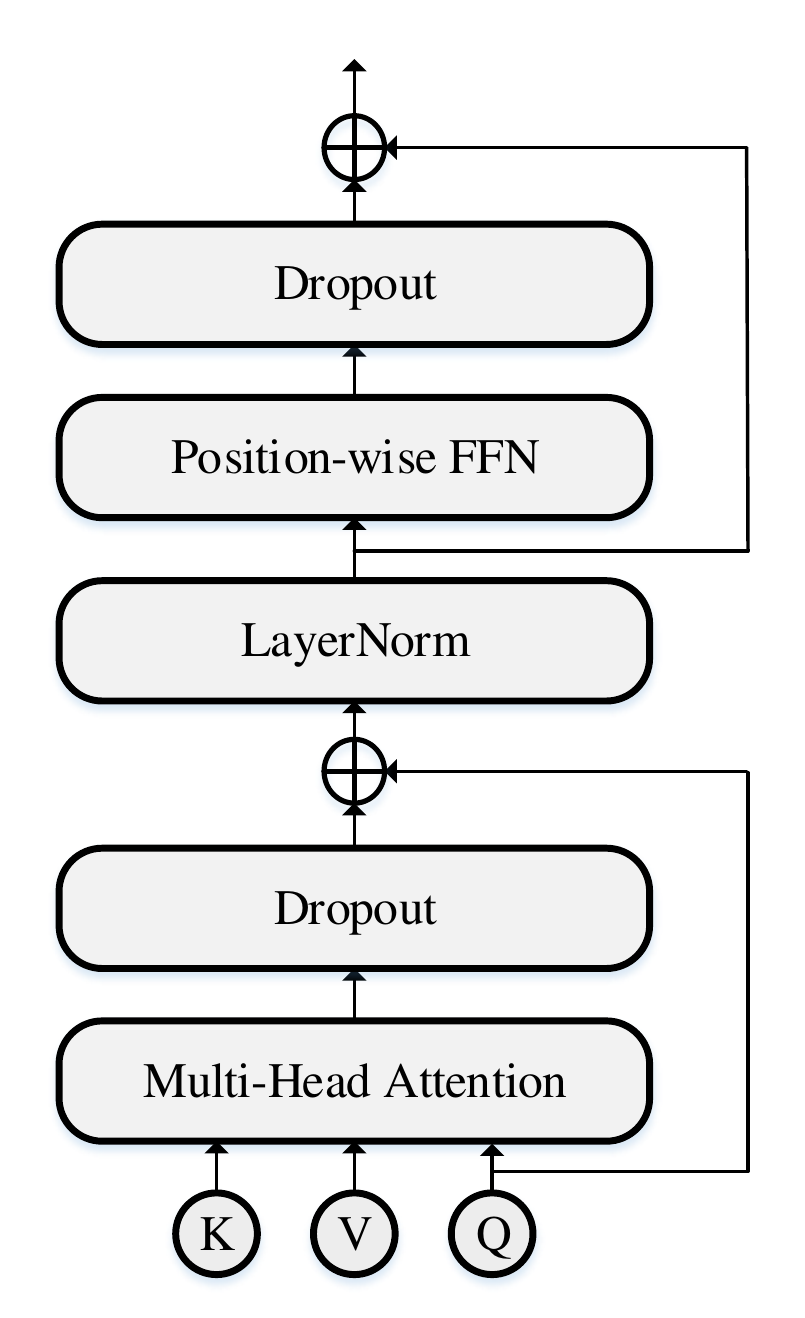} 
	\caption{An illustration of an attention block.} 
	\label{fig:block}
	\vspace{-10pt}
\end{figure}

\subsection{Attention Block}
The feedforward attention structure \cite{vaswani2017attention} captures the global relationship in a sequence. Different from recurrent neural networks which encode history context into latent vectors, the feedforward attention mechanism uses a weighted sum to fuse the input sequence. Because of its feedforward structure, it can be computed in parallel. In this work, we use the scaled dot-product attention and position-wise feedforward network as the basic submodule, following \cite{vaswani2017attention}. But we use ``pre-norm'' \cite{nguyen2019transformers} for stable training. The structure is shown in \autoref{fig:block}.

The scaled dot-product attention is computed by 
\begin{equation}
\label{eq:dot-product}
\text{Atten}(Q, K, V) = \text{Softmax}(\frac{QK^{T}}{\sqrt{D_k}})V,
\end{equation}
where $Q \in \mathbb{R}^{T_q \times D_{k}}$ denotes the queries, $K \in \mathbb{R}^{T_k \times D_{k}}$ denotes the keys, and $V \in \mathbb{R}^{T_k \times D_{v}}$ denotes values. As shown in \autoref{fig:attention}, the attention scores are computed with the dot products of queries and keys, then are normalized with softmax functions. The normalized attention scores will be sharp at some positions, and others are small. Then, by matrix multiplication, the values are fused to the corresponding position. This procedure can be seen as that the query queries keys and fetches out a corresponding value from values.

To make the attention scores various, it can be extended to a multi-head version:
\begin{equation}
\label{eq:multi}
\begin{split}
&\text{MHA}(Q, K, V) = \text{Concat}(h_1,\cdots,h_H)W^o, \\
&h_i = \text{Atten}(QW_i^q, KW_i^k, VW_i^v), i = 1,\cdots,H.
\end{split}
\end{equation}
The queries, keys, and values are transformed into subspaces with parameter matrices $W_i^q$, $W_i^k$, $W_i^v$, where $i$ is the index of a head. Then, the scaled dot-product attention is computed for the transformed inputs. At last, the outputs are concatenated together and multiplied with $W^o$. $h_i$ is one attention head. $H$ is the number of heads.

A position-wise feedforward neural network (FFN) transforms the output of the attention at each position: 
\begin{equation}
\label{eq:ffn}
\text{FFN}(u) = W_2 \text{Activate} (W_1 u + b_1) + b_2,
\end{equation}
where $u$ is a vector at one position, $W_1$, $W_2$, $b_1$, and $b_2$ are learnable parameters, ``$\text{Activate}$'' is a nonlinear activation function. In this work, gated linear units (GLUs) \cite{dauphin2017language} are used.

Residual connection \cite{he2016deep} and layer normalization (LN) \cite{ba2016layer} are also used. Different from \cite{vaswani2017attention}, we set the LN layer before the residual connection, following \cite{nguyen2019transformers} to make training stable and effective. The structure of the attention block is shown in \autoref{fig:block}.

\vspace{-10pt}
\subsection{Encoder}
The encoder extracts high-level representations from the acoustic feature sequence. We first use a two-layer convolutional neural network (CNN) to subsample the acoustic feature sequence. We set the stride on the time axis to $2$, so that the frame rate is reduced to $1/4$. Then the outputs of the CNN are flattened to a $T$-by-$D_m$ matrix, where $T$ is the length of the subsampled feature sequence, and $D_m$ is the dimensionality. Another purpose of the CNN is to capture the locality of the acoustic feature sequence.

Then, the encoder has a stack of $N_e$ attention blocks, as shown in \autoref{fig:arch}. The keys, queries, and values are all the same, so it is a self-attention mechanism. That is, the attention scores are obtained by computing dot-product between every two vectors of the inputs. Therefore, the long-term dependency is captured.

\vspace{-10pt}
\subsection{Position Dependent Summarizer}
The PDS module is the core of the proposed LASO model. The PDS module leverages queries, which depend on positions of the token sequence, to query the high-level representations from the encoder. It can be seen that the module ``summarizes'' the acoustic features, so we name it as ``summarizer''. As the result, it bridges the gap between the length of the acoustic feature sequence and the length of the token sequence. 

The PDS module consists of $N_s$ attention blocks, as shown in \autoref{fig:arch}. For the first block, the queries are position encodings. And the queries of the other blocks are the outputs of the previous block. Each query represents the token position in the token sequence. The keys and the values are all the outputs of the encoder. The length of $L$ of the position encoding sequence is pre-set by counting the length of utterances in the training set and add some tolerance. For example, if the maximum length of token sequences in the training set is 90, we can set this $L$ to 100. 

We use sinusoidal position encodings \cite{vaswani2017attention}:
\begin{equation}
\label{eq:pe}
\begin{split}
\text{pe}_{i, 2j} &= \text{sin}(i / 10000^{2j/D_m}),\\
\text{pe}_{i, 2j+1} &= \text{cos}(i / 10000^{2j/D_m}), \\
\end{split}
\end{equation}
where $i = 1, \cdots, L$ denotes the $i$-th position and $j$ is the index of an element. A benefit of using sinusoidal position encodings is that it would allow the model to easily learn relativity of positions. That is, the positional encoding of a fixed distance between two positions $k$ can be represented as a linear function of the two positions \cite{vaswani2017attention}.

\subsection{Decoder}
The decoder further refines the representations of the PDS module. Similar to the encoder, it is a self-attention module, which consists of $N_d$ attention blocks. It captures the relationships in the sequence, i.e., implicit language semantics, which is queried by the PDS. The inputs of the decoder are the outputs of the PDS. This decoder leverages the whole context of the utterance.

After the decoder, a linear transformation and a softmax function are used to compute the probability distribution over the token vocabulary. 

\begin{figure}[!t]\centering
	\includegraphics[width=1.0\columnwidth]{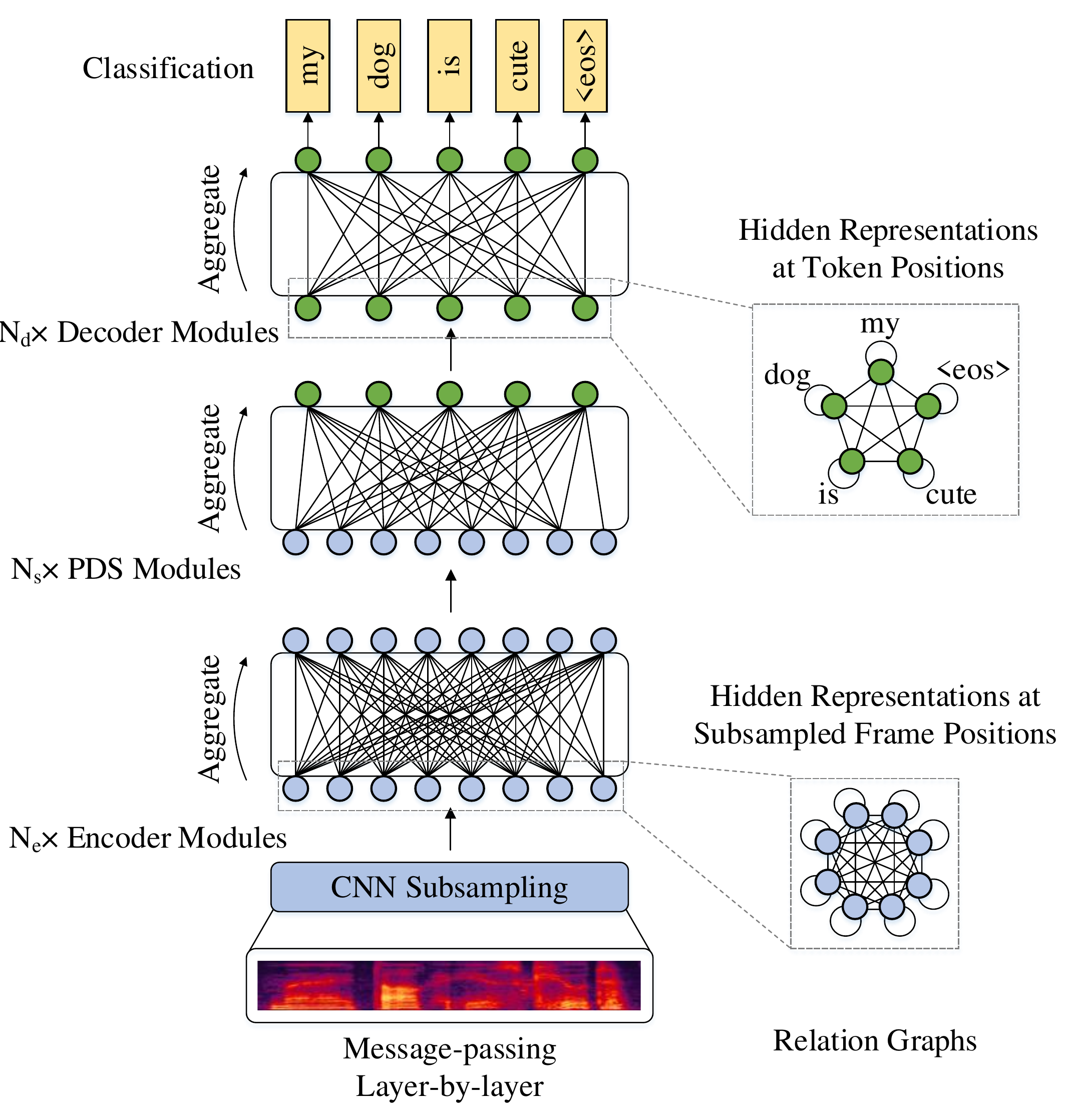} 
	\caption{An illustration of the LASO in a graph-based view. The left part shows that the messages are passed layer-by-layer. Each module (i.e. encoder module, PDS module, decoder module) aggregates the previous representations by the attention mechanism. The right part shows the relationships among representations in one layer.} 
	\label{fig:graph}
	\vspace{-10pt}
\end{figure}

\vspace{-10pt}
\subsection{The Graph-Based View of the Model}
We introduce a graph-based view of LASO to further illustrate the proposed model, as shown in \autoref{fig:graph}. This view can be divided into two aspects: a relation graph \cite{Jiaxuan2020graph} in one layer and message-passing of the forward procedure \cite{Justin2017message,guo2019star}.

For the aspect of the relations of the representations in one layer, the hidden representations can be seen as a fully connected graph with self-loops. Specifically, for the encoder which processes the acoustic features, the hidden representation at each position is related to all the hidden representations (including itself). Thus, the representation sequence can be viewed as a fully connected graph (the graph with blue nodes in the right part of \autoref{fig:graph}). With the PDS module, the representations are converted and are related to token positions. These representations correspond to each token position and consist of another fully connected graph (the graph with green nodes in the right part of \autoref{fig:graph}), i.e., one representation is related to all the representations in the sequence. The weights on the edges are computed with the attention mechanism. This is different from the autoregressive models. For an autoregressive model, the representation of a token is related to the previous representations, but not all the representations in the sequence.

For the aspect of the forward procedure, the forward propagation can be seen as a message-passing procedure in each module. Specifically, the outputs and the inputs of a module consist of a bipartite graph, as shown in the left part of \autoref{fig:graph}. The weights of the edges are attention scores. All representations in the inputted sequence are aggregated to the corresponding node of the output sequence by the weighing sum. Therefore, the messages in the previous relation graphs are passed to the next one in the forward propagation.

The graph-based view shows how LASO processes the acoustic feature sequence as a whole and achieves hidden representations at token positions. It also shows why the hidden representations at token positions are related to all the other hidden representations, which is different from the autoregressive model.

\subsection{Formulation}
The LASO model can be formulated as follows:
\begin{equation}
\label{eq:LASO}
\begin{split}
Z &= \text{Enc}(X),\\
q_i &= \text{Summarize}(Z, \text{pe}_i), \quad i = 1, 2, \cdots, L,\\
Q &= [q_1, \cdots, q_L]\\
P(y_i|X) &= \text{Dec}(Q), \quad i = 1, 2, \cdots, L,\\
\end{split}
\end{equation}
where $X=[x_1, \cdots, x_T]$ is the feature sequence, $Z$ denotes the high-level representations encoded with the encoder, and the probability over the vocabulary at each position $P(y_i|X)$ is computed with the PDS and the decoder. Function ``$\text{Summarize}$'' represents the PDS module, which attends the outputs of the encoder in terms of the positional encodings. Because the positional encodings, which are deterministic but not random, are a part of the whole model, they are not written in the probability expression.

\section{Learning}
\label{sec:learning}
In this section, we introduce the learning procedure of the LASO model.

\subsection{Maximum Likelihood Estimation}
We use \textit{maximum likelihood estimation} (MLE) criterion to train the parameters of the LASO model. We minimize the following negative log-likelihood (NLL) loss.
\begin{equation}
\label{eq:mle}
\text{NLL}(\theta) =  - \frac{1}{NL}\sum_{n=1}^{N} \sum_{i=1}^{L} \log P_{\theta}(y_i^{(n)}|X^{(n)}),
\end{equation}
where $X^{(n)}, Y^{(n)}$ is the $n$-th speech-text pair in the corpus. The total number of the pairs is $N$. $y_i^{(n)}$ is $i$-th token in $Y^{(n)}$. $L$ is the length of the token sequence, which is preset. If the length of a text data is shorter than $L$, \texttt{<eos>} is used to pad it, as shown in \autoref{fig:arch}. Thus, the length of the token sequence is estimated automatically. $\theta$ denotes trainable parameters of the model. 

This training procedure is end-to-end and does not depend on any frame-level label generated by some GMM-HMM system.

\begin{figure}[!t]\centering
	\includegraphics[width=0.9\columnwidth]{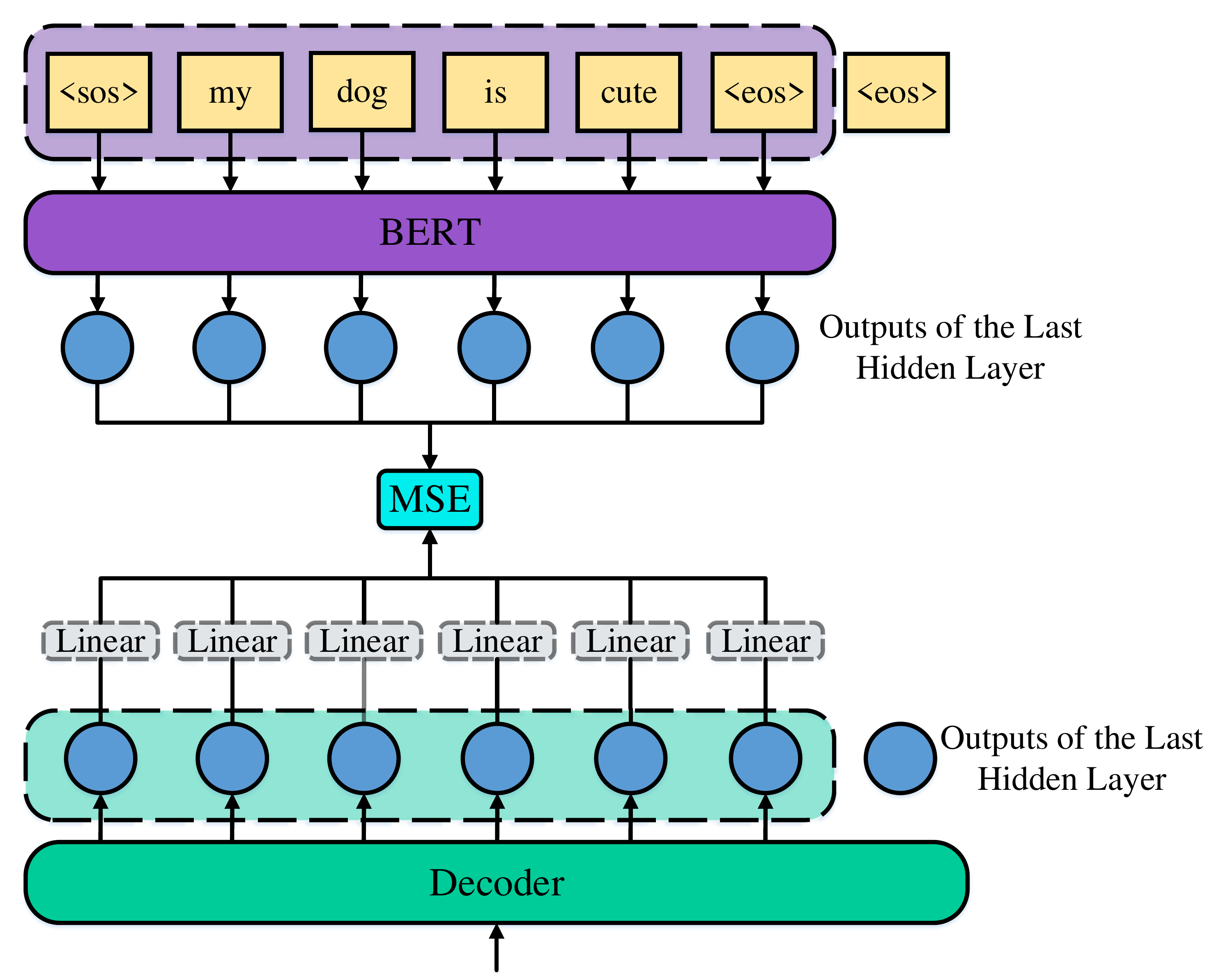} 
	\caption{An illustration of semantic refinement from BERT. The valid part of the token sequence is inputted into BERT. And the MSE loss between the last hidden layers of decoder and BERT is minimized. The optional linear transformation is used when the dimensionalities of the decoder and BERT are different.} 
	\label{fig:bert}
\end{figure}

\subsection{Semantic Refinement from BERT}
To further improve the performance, we use teacher-student learning \cite{2006model,li2014learning,hinton2015distilling,romero2014fitnets} to refine knowledge from pre-trained LM. BERT, a kind of denoising autoencoder LM trained on very large-scale text, has shown the powerful ability of language modeling and achieved state-of-the-art performance on many NLP tasks \cite{devlin2018bert}. Inspired by our previous paper \cite{bai2019learn}, we transfer the knowledge from BERT to the LASO model. Another potential advantage of using BERT as the teacher model is that both our proposed LASO and BERT are bidirectional models, i.e., the model predicts a token using both the left context and the right context. 

The basic idea is that the BERT can provide a good semantic representation for each token. And we consider the outputs of the decoder also provide token-level representation. Therefore, we make the outputs of the decoder approximate the BERT. We minimize the mean squared error (MSE) between their last hidden layers \cite{romero2014fitnets}, as shown in \autoref{fig:bert}. To match the training procedure of BERT, we add a \texttt{<sos>} token at the head of the token sequence. 

We input the token sequence into BERT model to fetch out the outputs of the last hidden layer. Note that we only input the valid part of the token sequence, but not the padding token \texttt{<eos>} at the tail. \texttt{<sos>} and \texttt{<eos>} are converted to \texttt{[CLS]} and \texttt{[SEP]}, which are two special tokens added at the head and the tail of a sentence in BERT \cite{devlin2018bert}.

We also use the valid part of the outputs of the decoder, i.e., the outputs corresponding to the subsequence from \texttt{<sos>} to the first \texttt{<eos>}. Then, we compute the MSE:
\begin{equation}
\label{eq:bert}
\text{MSE}(\theta) = \frac{1}{N} \sum_{n=1}^{N} \frac{1}{L_{v}^{(n)}} \sum_{i=1}^{L_{v}^{(n)}}\sum_{d=1}^{D_B}(h_{i, d}^{(n)} - s_{i, d}^{(n)})^2,
\end{equation}
where $h_{i, d}^{(n)}$ is the $d$-th element of the $i$-th vector in the decoder output sequence for the $n$-th data, and $s_{i, d}^{(n)}$ is the $d$-th element of the $i$-th vector in the BERT output sequence for the $n$-th data, $D_B$ is the dimensionality of BERT output, $L_{v}^{(n)}$ is the valid length of $n$-th data, and $N$ is the total number of the data. $\theta$ represents all parameters of the model.

If the dimensionality of the decoder is different from BERT, we can simply use an optional linear transformation to make the dimensionality matching, as shown in \autoref{fig:bert}.

Another benefit of transferring knowledge with the hidden layer rather than probability is that it makes the construction of vocabulary more flexible. We can use only a part vocabulary of BERT rather than all tokens.

At last, we combine the NLL loss and the MSE loss as the final loss:
\begin{equation}
\label{eq:final}
L(\theta) = \text{NLL}(\theta) + \lambda \text{MSE}(\theta),
\end{equation}
where $\lambda$ is a coefficient to balance the values of the two losses. The typical value is $0.005$.

The BERT model is only used at the training stage. It does not add any extra complexity during inference.

\section{Inference}
\label{sec:inference}
The inference of LASO is simple. We just select the most likely token at each position:
\begin{equation}
\label{eq:infer}
\hat{y_i} = \arg\max_{y_i} P(y_i|X). \quad i=1,\cdots,L,
\end{equation}
Then, the special tokens \texttt{<eos>} (or including \texttt{<sos>} if the model is trained with BERT) are removed. Note that the positional encodings ($[\text{pe}_i; \cdots; \text{pe}_L]$) are a part of the whole model so that they are inputted into the model as a whole.

This inference procedure does not depend on beam-search, so multi-pass forward propagation is not needed. Thus, the inference time cost is much reduced.

\begin{table*}[!t]
	\caption{The Description of the Datasets}
	\centering
	\begin{tabular}{clccccccccc}
		\toprule
		\multicolumn{2}{l}{\multirow{2}{*}{}}          & \multirow{2}{*}{\#Utter.} & \multirow{2}{*}{\#Hours} & \multirow{2}{*}{\#Speaker} & \multicolumn{3}{c}{Duration (Sec.)} & \multicolumn{3}{c}{\#Token Per Sentence} \\ \cmidrule(l){6-11} 
		\multicolumn{2}{l}{}                           &                           &                          &                            & Min.       & Max.       & Avg.      & Min.         & Max.        & Avg.        \\ \midrule
		\multirow{3}{*}{AISHELL-1} & Training          & 120,098                   & 150                      & 340                        & 1.2        & 14.5       & 4.5       & 1.0          & 44.0        & 14.4        \\
		& Dev.              & 14,326                    & 18                       & 40                         & 1.6        & 12.5       & 4.5       & 3.0          & 35.0        & 14.3        \\
		& Test              & 7,176                     & 10                       & 20                         & 1.9        & 14.7       & 5.0       & 3.0          & 21.0        & 14.0        \\ \midrule
		\multirow{7}{*}{AISHELL-2} & Training (iPhone) & 1,009,223                 & 1000                     & 1347                       & 0.5        & 19.3       & 3.6       & 1.0          & 53.0        & 10.9        \\
		& Dev. (iPhone)     & 2,500                     & 2                        & 5                          & 1.1        & 9.4        & 2.9       & 1.0          & 25.0        & 9.9         \\
		& Dev. (Android)    & 2,500                     & 2                        & 5                          & 1.1        & 9.4        & 2.9       & 1.0          & 25.0        & 9.9         \\
		& Dev. (HiFi Mic.)  & 2,500                     & 2                        & 5                          & 1.1        & 9.4        & 2.9       & 1.0          & 25.0        & 9.9         \\
		& Test (iPhone)     & 5,000                     & 4                        & 10                         & 1.1        & 8.5        & 2.9       & 1.0          & 25.0        & 9.9         \\
		& Test (Android)    & 5,000                     & 4                        & 10                         & 1.1        & 8.5        & 2.9       & 1.0          & 25.0        & 9.9         \\
		& Test (HiFi Mic.)  & 5,000                     & 4                        & 10                         & 1.1        & 8.5        & 2.9       & 1.0          & 25.0        & 9.9         \\ \bottomrule
	\end{tabular}
	\label{tab:data}
	\vspace{-10pt}
\end{table*}

\section{Related Work}
\label{sec:rel}
In this section, we review and compare the previously related work in two main aspects. One is the non-autoregressive AED model. And the other is the utilization of LMs for AED models.

\subsection{Non-Autoregressive AED Models}
Non-autoregressive AED models are first used in machine translation (MT). Gu \textit{et al.} first proposed non-autoregressive machine translation and introduced fertility to tack the multimodality problem \cite{gu2017non}. Auxiliary regularization \cite{wang2019non} and enhanced decoder input \cite{guo2019non} are proposed to improve the performance. Lee \textit{et al.} proposed an iterative refinement algorithm for MT \cite{lee2018deterministic}. MA \textit{et al.} proposed a flow model for sequence generation \cite{ma2019flowseq}. These models showed promising results on both performance and efficiency. However, these models are used for MT but not ASR. Speech signal has its specific property, for example, the monotonic alignment with the token sequence, and the implicit language semantics in speech. This motivated us to propose a simpler non-autoregressive AED model for ASR. The non-autoregressive transformer, which completes the masked tokens iteratively, is proposed for ASR \cite{chen2019non}. MASK-CTC uses decoding results as the initialization and refines the results with a masked language model \cite{Higuchi2020}. Different from the previous work, we propose a non-autoregressive model LASO, which forward propagates for one-pass. We reformulate speech recognition as a position-wise classification problem. We propose the PDS to extract token-level representation from speech. The PDS bridges the length gap between the speech and the token sequence.

\subsection{The Utilization of LMs}
Fusion methods, such as shallow fusion, deep fusion, and cold fusion, integrate external LMs to improve the performance \cite{gulcehre2015on,sriram2018cold}. However, these methods add extra complexity during inference. And these methods can only use unidirectional LMs, so recent powerful bidirectional LMs such as ELMo \cite{peters2018deep}, BERT \cite{devlin2018bert}, are not applicable. BERT was used to rescore the n-best results \cite{shin2019effective}. However, rescoring increases computation during inference. And it does not use the representation ability of BERT. Bai et al. proposed LST approach to transfer knowledge from an LM to an ARM with teacher-student learning \cite{bai2019learn}. \cite{futami2020distilling} distilled knowledge from BERT to an ARM. In this paper, we transfer the knowledge from BERT to improve the LASO model. LASO and BERT both capture the global token-level relationship. With this method, the LASO model can benefit from the representation ability of BERT but does not add any extra computation during inference.

\subsection{Cross-Modal Semantic Alignment}
The semantic refinement from BERT is also related to cross-modal semantic alignment, i.e., aligning the spaces of LASO and BERT. The concept of cross-modal semantic alignment is first used in cross-modal retrieval \cite{lavrenko2003model,jeon2003automatic,jiang2017deep,fan2017cross,zhen2019deep,yang2020learning}. Especially the recent deep learning work learn a shared semantic space between the images and the text-based on neural networks \cite{fan2017cross,jiang2017deep,zhen2019deep,yang2020learning}. The basic idea of these work is to map the features from different modalities so that the system can easily compute their similarities. Recent ASR-free approach to text query-based keyword search from a speech also based on this idea, i.e., learning a shared semantic space between text query and speech \cite{audhkhasi2017end}. The motivation of our proposed semantic refinement from BERT is also to align semantic of LASO and BERT. However, different from the retrieval work, we do not train the models of the two modalities but only train the LASO model. Namely, the BERT model, which has been confirmed as a powerful language model, is an auxiliary model to train the LASO model and is not used during inference. In addition, this work is in a cross-model knowledge transferring setting, i.e., transferring knowledge from text-modal model BERT to speech-modal model LASO\footnote{Different from AED models, the inputs of LASO are only speech features but not text embeddings. Therefore, it is a unimodal model.}.

\section{Experimental Setup}
\label{sec:setup}
In this section, we introduce the used datasets and the experimental setup. All the experiments are implemented with deep learning toolkit PyTorch \cite{Paszke2019pytorch} with python programming language.

\subsection{Datasets}
We conduct experiments on public Chinese speech datasets AISHELL-1\footnote{http://www.openslr.org/33/} \cite{bu2017aishell} and AISHELL-2\footnote{http://www.aishelltech.com/aishell\_2} \cite{du2018aishell}. These two datasets have different scales of data so that we can evaluate the generalization on both a small dataset and a large dataset. The reason to select these two datasets is that character-based tokenization of the officially pre-trained Google's BERT for Chinese is proper for the experiment of semantic refinement from BERT \footnote{We did some initial experiments on English dataset LibriSpeech. However, the tokenization method of English BERT makes the vocabulary very large (27563 English word pieces) so that it is not easy for the model to converge. We noticed similar phenomena of MASK-CTC \cite{Higuchi2020} on Latin-alphabet-based experiments, while it performs very well in the Japanese task. In the future, we will try to select proper modeling units for Latin-alphabet-based data and train our own BERT models. We provide discussion on English dataset LibriSpeech in the supplemental materials.}.

AISHELL-1 contains 178 hours of Mandarin speech. The speech is recorded by 400 speakers. All audio is recorded with high fidelity microphones in 44.1 kHz, then subsampled to 16 kHz. The content of the datasets covers 5 domains including ``Finance'', ``Science and Technology'', ``Sports'', ``Entertainments'', and ``News''.

AISHELL-2 contains about 1000 hours of Mandarin Speech for training. The training set is recorded by 1991 speakers with iPhone smartphones. The content covers voice commands, digital sequence, places of interest, entertainment, finance, technology, sports, English spellings, and free speaking without specific topics. The development sets and the test sets are recorded with different equipment to evaluate generalization for different equipment.

The details of the two datasets are shown in \autoref{tab:data}.

\begin{table}[!t]
	\caption{The Symbols of the Hyper-parameters of the Architecture}
	\centering
	\begin{tabular}{@{}c|l@{}}
		\toprule
		Symbol     & \multicolumn{1}{c}{Description}                               \\ \midrule
		$D_m$      & The dimensionality of the inputs of the multi-head attention. \\
		$D_{in}$   & The inner dimensionality of the position-wise FFN.            \\
		Activation & The type of activation function of the position-wise FFN.     \\
		\#Enc.     & The number of blocks of the encoder.                          \\
		\#PDS      & The number of blocks of PDS.                                  \\
		\#Dec.     & The number of blocks of the decoder.                          \\ \bottomrule
	\end{tabular}
	\label{tab:symbol}
	\vspace{-10pt}
\end{table}

\subsection{Setup}
\textbf{Basic settings}. We first evaluate the models on the small-scale (150 hours) dataset AISHELL-1. Then we extend the experiments to the large-scale (1000 hours) dataset AISHELL-2. We use $80$-dimension Mel-filter bank features (FBANK) as the inputs, which are extracted every 10ms with 25ms of frame length. For AISHELL-1, the token vocabulary contains $4231$ characters in the training set and three special symbols, i.e., "\texttt{<sos>}" for the start of the sentence, "\texttt{<unk>}" for unseen characters, and "\texttt{<eos>}" as the filler of the tail of a token sequence. The vocabulary size for AISHELL-2 is $5252$. 

\textbf{Baseline settings}. We build an autoregressive AED model Speech-Transformer \cite{dong2018speech,zhou2018syllable} and a non-autoregressive SAN-CTC model \cite{Salazar2019self} as two baselines. The two models use the same settings as LASO, including the above-mentioned features, vocabulary, and basic attention blocks (\autoref{fig:block}). Both two models use the whole input signal, which is the same with LASO. Because we would like to compare the performance of SAN-CTC in the end-to-end non-autoregressive setting, the decoding process is greedy, i.e., directly select the most likely token and remove the blanks, without beam-search on an N-gram LM based searching graph.

For Speech-Transformer, both the encoder and the decoder have 6 attention layers. The model dimensionality is 512. The number of heads of the attention is 8. The dimensionality of the intermediate FFN is 2048. And the activation function is GLU. It also uses the same CNN subsampling layer as the LASO models, as shown in \autoref{fig:arch}. We refer to this model as \texttt{Transformer}. For SAN-CTC, because it only has an encoder part, we set the number of the attention layer to 12. The model size is comparable to \texttt{Transformer} and other models. We refer to this model as \texttt{SAN-CTC}.

\textbf{LASO settings}. We compare different architectures of LASO. The symbols of network configuration are shown in \autoref{tab:symbol}. The basic attention blocks are shown in \autoref{fig:block}, which are the same as the two baselines. The lengths of positional encodings of the PDS module are set to $60$ ($L$ in \autoref{eq:LASO}).

\textbf{Training settings}. We use Adam algorithm \cite{kingma2014adam} to optimize the models. We use the warm-up learning rate schedule \cite{vaswani2017attention}:
\begin{equation}
\alpha = D^{-0.5}  \cdot \text{min} (step^{-0.5}, step \cdot warmup^{-1.5}). 
\end{equation}
The warm-up step is set to 12000. The dropout rate is set to 0.1. Each batch contains about 100 seconds of speech, and we accumulate gradients of 12 steps to simulate a big batch \cite{ott2018scaling} for stabilizing training. We train the models until they converge. The typical number of epochs for LASO is 130. And the typical number of epochs for Speech-Transformer and SAN-CTC is 80.

We use SpecAugment \cite{park2019specaugment} for data augmentation. The frequency masking width is 27. The time masking width is 40. Both frequency masking and time masking are employed twice. But we do not use time warping. We leverage label smoothing with 0.1 for over-confidence problems during training. We average parameters of the models which are saved at the last 10 epochs as the final model.

We use the Google's pre-trained Chinese BERT model\footnote{https://storage.googleapis.com/bert\_models/2018\_11\_03/chinese\_L-12\_H-768\_A-12.zip} for semantic refinement. This model has 12 transformer layers. The model dimensionality is 768. The total number of parameters is 110M. The vocabulary of the BERT model contains 21128 tokens. During training, the coefficient $\lambda$ in \autoref{eq:final} is set to $0.005$. 

\begin{table*}[!t]
	\caption{The Character Error Rates on AISHELL-1 with Different Hyper-parameters}
	\centering
	\begin{tabular}{cccccccccccc}
		\toprule
		\multirow{2}{*}{Model} &
		\multirow{2}{*}{Dm} &
		\multirow{2}{*}{Din} &
		\multirow{2}{*}{Activation} &
		\multirow{2}{*}{\#Enc.} &
		\multirow{2}{*}{\#PDS} &
		\multirow{2}{*}{\#Dec.} &
		\multirow{2}{*}{\begin{tabular}[c]{@{}c@{}}Model\\ size\end{tabular}} &
		\multicolumn{2}{c}{w/o BERT} &
		\multicolumn{2}{c}{w/ BERT} \\ \cmidrule(l){9-12} 
		&     &      &      &   &   &   &        & CERs on Dev. & CERs on Test & CERs on Dev. & CERs on Test \\ \midrule
		1  & 256 & 2048 & ReLU & 4 & 1 & 4 & 15.9M  & 7.9          & 8.8          & 7.7          & 8.6          \\
		2  & 256 & 2048 & GLU  & 4 & 1 & 4 & 20.6M  & 7.1          & 8.1          & 7.0          & 7.8          \\ \midrule
		3  & 256 & 2048 & ReLU & 6 & 1 & 6 & 21.2M  & 7.2          & 8.2          & 7.1          & 8.0          \\
		4  & 256 & 2048 & GLU  & 6 & 1 & 6 & 28.0M  & 6.6          & 7.5          & 6.6          & 7.4          \\ \midrule
		5  & 256 & 2048 & ReLU & 6 & 2 & 6 & 22.8M  & 7.4          & 8.3          & 7.3          & 8.1          \\
		6  & 256 & 2048 & GLU  & 6 & 2 & 6 & 30.1M  & 6.6          & 7.5          & 6.3          & 7.1          \\ \midrule
		7  & 256 & 2048 & ReLU & 8 & 2 & 6 & 25.4M  & 7.3          & 8.4          & 7.3          & 8.1          \\
		8  & 256 & 2048 & GLU  & 8 & 2 & 6 & 33.8M  & 6.7          & 7.5          & 6.5          & 7.4          \\ \midrule
		9  & 512 & 2048 & ReLU & 4 & 1 & 4 & 37.0M  & 7.0          & 7.8          & 6.6          & 7.4          \\
		10 & 512 & 2048 & GLU  & 4 & 1 & 4 & 46.5M  & 6.4          & 7.4          & 5.9          & 6.7          \\ \midrule
		11 & 512 & 2048 & ReLU & 6 & 1 & 6 & 49.6M  & 6.6          & 7.5          & 6.0          & 6.7          \\
		12 & 512 & 2048 & GLU  & 6 & 1 & 6 & 63.3M  & 6.2          & 7.0          & 5.4          & 6.2          \\ \midrule
		13 & 512 & 2048 & ReLU & 6 & 2 & 6 & 53.9M  & 6.6          & 7.5          & 6.6          & 7.5          \\
		14 & 512 & 2048 & GLU  & 6 & 2 & 6 & 68.6M  & 6.1          & 6.9          & 5.4          & 6.1          \\ \midrule
		15 & 512 & 2048 & ReLU & 8 & 2 & 6 & 60.2M  & 6.6          & 7.5          & 6.2          & 6.9          \\
		16 & 512 & 2048 & GLU  & 8 & 2 & 6 & 80.0M  & 5.9          & 6.6          & 5.2          & 5.8          \\ \midrule
		17 & 768 & 2048 & ReLU & 6 & 1 & 6 & 85.5M  & 6.5          & 7.3          & 5.9          & 6.6          \\
		18 & 768 & 2048 & GLU  & 6 & 1 & 6 & 105.8M & 5.9          & 6.9          & 5.3          & 6.1          \\ \bottomrule
	\end{tabular}
	\label{tab:hyper}
	\vspace{-10pt}
\end{table*}

\textbf{Performance evaluation}. For performance evaluation, we use the standard edit-distance based error rate, i.e. character error rate (CER). For speed evaluation, we use both real-time factor (RTF) and average processing time (APT)\cite{gu2017non}. RTF is a standard metric to evaluate the processing time cost of an ASR system. It is the average time cost to process one-second speech:
\begin{equation}
\text{RTF} = \frac{\text{Total Processing Time}}{\text{Total Duration}}.
\end{equation}
This metric is dimensionless and independent of utterance duration. To consider the impact to the processing time of the utterance duration, we compute APT:
\begin{equation}
\text{APT} = \frac{\text{Total Processing Time}}{\text{Total Number of Utterance}}.
\end{equation}
The unit of APT is second.

The reason to use ATP is to show the efficiency to process one utterance. It considers the waiting time of a user and is suitable for both online applications (such as speech interaction systems) and offline applications (such as voice document transcription). Specifically, for online applications, the waiting time of a user is ``the time between the stop-point at which the user stops speaking and the appearance on the screen of the ASR result''. And for offline applications, the waiting time of a user is ``the time between the point at which the user inputs the utterance and the appearance on the screen of the ASR result''. These metrics ignore the uncontrollable factors of a speech engineer, e.g., data transmission speed on the Internet. 

RTF may give an \textit{overestimated sense} of whole-utterance ASR systems (AED, Speech-Transformer, or bidirectional AM based hybrid models), because the whole-utterance ASR systems starts processing speech after receiving the whole utterance, unlike some streaming models. For streaming models, the waiting time of a user of an online application can be estimated as 
\begin{equation}
\text{RTF} \times \text{the length of the last data package}.
\end{equation}
However, for whole-utterance ASR systems, the length of whole-utterance is needed to be considered. To address this issue, we compute APT, which directly average the processing time cost of an utterance. We compute these two practical values on commonly used deep learning devices (GPUs) to show the time cost of the ASR systems. We also mention that the time cost of feature extraction is included in our experimental implementation. 
\vspace{-5pt}

\section{Experimental Results}
\label{sec:res}

In this section, we introduce the experimental results.
\vspace{-5pt}

\subsection{Comparing Model Architectures on AISHELL-1}

\begin{table}[!t]
	\caption{Comparisons with Baselines on AISHELL-1}
	\centering
	\begin{threeparttable}
		\resizebox{\linewidth}{!}{	
			\begin{tabular}{lcccc}
				\toprule
				\multicolumn{1}{c}{\multirow{2}{*}{Model}} & \multirow{2}{*}{\#Param.} & \multicolumn{2}{c}{CERs} & \multirow{2}{*}{RTF / APT} \\ \cmidrule(lr){3-4}
				\multicolumn{1}{c}{}      &       & Dev. & Test &            \\ \midrule
				KALDI(nnet3) * $\dagger$ $\ddagger$           & -     & -    & 8.6  & -          \\
				KALDI(chain) * $\dagger$ $\ddagger$           & -     & -    & 7.4  & -          \\
				LAS \cite{sun2019adversarial}                 & -     & 9.4  & 10.6 & -          \\
				ESPNet (Transformer) $\dagger$ $\ddagger$ \cite{karita2019comparative}     & -     & 6.0    & 6.7  & -          \\
				A-FMLM   \cite{chen2019non}                 & -     & 6.2  & 6.7  & -          \\
				Fan et al. (Transformer) \cite{fan2019unsupervised}  & -     & -    & 6.7  & -          \\
				AGS CTC $\ddagger$ \cite{ding2020ctc}  & -     & 7.0  & 7.9  & -          \\ \midrule
				\texttt{Transformer} (baseline1)        & 67.5M & 6.1  & 6.6  & 0.19 / 961ms   \\
				\texttt{SAN-CTC} (baseline2) 		   & 56.4M & 7.2  & 7.8  & 0.0033 / 16ms  \\
				\texttt{LASO-small}                & 20.6M & 7.1  & 8.1  & 0.0027 / 13ms  \\
				\texttt{LASO-small} w/ BERT        & 20.6M & 7.0  & 7.8  & 0.0027 / 13ms  \\
				\texttt{LASO-middle}               & 63.3M & 6.2  & 7.0  & 0.0035 / 17ms  \\
				\texttt{LASO-middle} w/ BERT       & 63.3M & 5.4  & 6.2  & 0.0035 / 17ms  \\
				\texttt{LASO-big}                  & 80.0M & 5.9  & 6.6  & 0.0040 / 20ms  \\
				\texttt{LASO-big} w/ BERT          & 80.0M & \textbf{5.2}  & \textbf{5.8}  & 0.0040 / 20ms  \\ \bottomrule
			\end{tabular}
		}
		\begin{tablenotes}	
			\footnotesize
			\item[*] from the KALDI official repository\footnotemark.
			\item[$\dagger$] with speed perturbation based data augmentation.
			\item[$\ddagger$] with an extra language model at the inference stage.
		\end{tablenotes}
	\end{threeparttable}
	\label{tab:compare_aishell1}
	\vspace{-10pt}
\end{table}

\begin{table*}[!t]
	\caption{Comparisons with Baselines on AISHELL-2}
	\centering
	\begin{threeparttable}
		\begin{tabular}{@{}lccccccccc@{}}
			\toprule
			\multicolumn{1}{c}{\multirow{3}{*}{Model}} & \multirow{3}{*}{\#Param.} & \multicolumn{8}{c}{CERs}                                                                                                                          \\ \cmidrule(l){3-10} 
			\multicolumn{1}{c}{}                       &                          & Dev.         & Dev.         & Dev.         & \multicolumn{1}{l}{Dev.}   & Test         & Test         & Test         & \multicolumn{1}{l}{Test}   \\
			\multicolumn{1}{c}{}                       &                          & (iPhone)     & (Android)    & (HiFi Mic.)  & \multicolumn{1}{l}{(Avg.)} & (iPhone)     & (Android)    & (HiFi Mic.)  & \multicolumn{1}{l}{(Avg.)} \\ \midrule
			KALDI (chain) \cite{du2018aishell} $\dagger$ $\ddagger$      & -                        & 9.1          & 10.4         & 11.8         & 10.4                       & 8.8          & 9.6          & 10.9         & 9.8                        \\
			LAS \cite{sun2019adversarial}     & -                        & -            & -            & -            & -                          & 9.2          & 9.7          & 10.3         & 9.7                        \\
			ESPNet (transformer) * $\dagger$ $\ddagger$ & -                        & -            & -            & -            & -                          & 7.5          & 8.9          & 8.6          & 8.3                        \\ \midrule
			\texttt{Transformer} (baseline1)               & 67.5M                    & 6.4          & 7.2          & 7.7          & 7.1                        & 7.1          & 8.0          & 8.2          & 7.8                        \\
			\texttt{SAN-CTC} (baseline2)                   & 56.4M                    & 8.3          & 8.9          & 8.8          & 8.6                        & 8.0          & 9.0          & 8.9          & 8.7                        \\
			\texttt{LASO-small}                        & 20.6M                    & 8.2          & 9.5          & 9.7          & 9.1                        & 8.5          & 9.5          & 9.5          & 9.2                        \\
			\texttt{LASO-small} w/ BERT                & 20.6M                    & 8.9          & 10.0         & 10.3         & 9.7                        & 8.8          & 9.8          & 10.5         & 9.7                        \\
			\texttt{LASO-middle}                       & 63.3M                    & 6.6          & 7.5          & 7.6          & 7.2                        & 6.8          & 7.4          & 7.3          & 7.2                        \\
			\texttt{LASO-middle} w/ BERT               & 63.3M                    & 6.5          & 7.2          & 7.4          & 7.0                        & 6.6          & \textbf{7.2} & \textbf{7.1} & 7.0                        \\
			\texttt{LASO-large}                        & 80.0M                    & 6.4          & 7.3          & 7.3          & 7.0                        & 6.7          & 7.4          & 7.4          & 7.1                        \\
			\texttt{LASO-large} w/ BERT                & 80.0M                    & \textbf{6.2} & \textbf{7.2} & \textbf{7.3} & \textbf{6.9}               & \textbf{6.5} & \textbf{7.2} & \textbf{7.1} & \textbf{6.9}               \\ \bottomrule
		\end{tabular}
		\begin{tablenotes}	
			\footnotesize
			\item[*] from the ESPnet official repository\footnotemark.
			\item[$\dagger$] with speed perturbation based data augmentation.
			\item[$\ddagger$] with an extra language model at the inference stage.
		\end{tablenotes}
	\end{threeparttable}
	\label{tab:compare_aishell2}
	\vspace{-10pt}
\end{table*}

First, we comparing the performance of the different model configurations. \autoref{tab:hyper} shows the results of the different configurations. From \autoref{tab:hyper}, we conclude statements as follows.

\begin{enumerate}
	\item $D_{m}$: A large number of the dimensionality of the attention block makes the model have more powerful representation ability and achieve better performance.
	\item Activation: We find that using GLU as the activation function is more effective than using ReLU consistently. The GLU will introduce more parameters to the model. However, comparing the deeper model with ReLU and the shallower model with GLU, whose model sizes are relatively similar, the model with GLU can achieve better performance.
	\item \#Enc. and \#Dec.: Increasing the number of the blocks of the encoder and the decoder improves the performance of the model.
	\item \#PDS: We compare the model with one-layer PDS and the one with two-layer PDS. We find the difference between the two is not very significant.
	\item Semantic Refinement from BERT: We can see that with semantic refinement from BERT, the performance improvements are significant for the $D_{m}=512$. However, for the models with $D_{m}=256$, the performance improvements are subtle. We analyze that it is because the discrepancy of the dimensionality of the LASO and BERT is too large so that it is hard for the LASO models to learn the representations of BERT.   
\end{enumerate}

\footnotetext{https://github.com/kaldi-asr/kaldi/blob/master/egs/aishell/s5/RESULTS}

In the rest of the paper, we refer to model 2, model 12, and model 16 in \autoref{tab:hyper} as \texttt{LASO-small}, \texttt{LASO-middle}, and \texttt{LASO-big}, respectively, to compare performance with previous models.

\subsection{Comparisons with Other Methods on AISHELL-1}
We compare our proposed LASO with baseline autoregressive model \texttt{Transformer} and non-autoregressive CTC model \texttt{SAN-CTC}. We also compare it with previous work. The results are shown in \autoref{tab:compare_aishell1}. We can see that our proposed LASO models achieve competitive performance, compared with the hybrid models, the CTC models, and autoregressive transformer models. In particular, with semantic refinement from BERT, \texttt{LASO-middle} and 
\texttt{LASO-big} outperform the autoregressive Transformer model. Moreover, the inference speed of the non-autoregressive models is much faster than the autoregressive models.

We implemented a competitive baseline autoregressive model \texttt{Transformer}, which achieves a CER of 6.6\% on the test set of AISHELL-1. The performance of \texttt{LASO-middle}, which has a similar scale of the number of the parameters with \texttt{Transformer}, is comparable to \texttt{Transformer}. And the bigger model \texttt{LASO-big} outperforms \texttt{Transformer}. Both \texttt{LASO-middle} and \texttt{LASO-big} outperform the CTC based non-autoregressive model \texttt{SAN-CTC}.  

\footnotetext{https://github.com/espnet/espnet/blob/master/egs/aishell2/asr1/RESULTS.md}

\autoref{tab:compare_aishell1} also lists RTF and APT of the models. RTF is the ratio of the total inference time to the total duration of the test set. APT is the averaged time cost for decoding one utterance (including the time of feature extraction) on the test set. The inference is done utterance by utterance on an NVIDIA RTX 2080Ti GPU. We can see that the inference speed of the non-autoregressive models is much faster than the autoregressive transformer model \texttt{Transformer}. The APT of the non-autoregressive models is reduced by $50\times$. And we can see that even the model size of \texttt{LASO-big} is larger than \texttt{Transformer}, the APT of \texttt{LASO-big} is much reduced. With the non-autoregressive and the feedforward structure, the inference of the model can be implemented very efficiently.

The inference speed of the baseline non-autoregressive model \texttt{SAN-CTC} is also very fast. However, compared with the similar scale LASO models, the performance is degraded. We analyze that the LASO models use the decoder, which plays the role of an autoencoder LM, to capture language semantics more effectively. In addition, for CTC models, the blank symbols inserted in the token sequence may influence the model to capture language semantics.

For \texttt{LASO-middle} and \texttt{LASO-big}, semantic refinement from BERT much reduces CERs by 11\% to 12\% on the test set. This demonstrates that teacher-student learning with BERT can improve the ability of the LASO to capture the language semantics. However, for the small-size model \texttt{LASO-small} semantic refinement from BERT does not improve the performance. We analyze that the dimensionality of the representations of \texttt{LASO-small} and BERT are very different (256 vs. 768), which influences the LASO model to learn knowledge from the BERT model.

\begin{figure*}[!t]
	\centering
	\subfloat[The attention scores of the first head of the last encoder layer.]
	{   \label{fig:enc_self_att_layer5_head0}
		\includegraphics[width=0.21\linewidth]{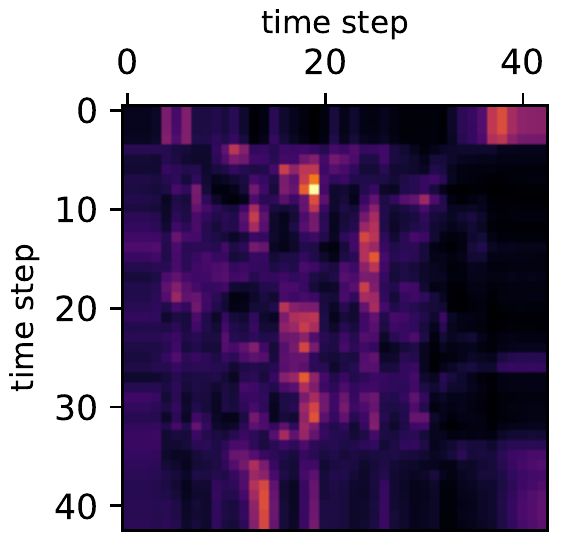}
	}		
	\quad
	\subfloat[The attention scores of the second head of the last encoder layer.]
	{   \label{fig:enc_self_att_layer5_head1}
		\includegraphics[width=0.21\linewidth]{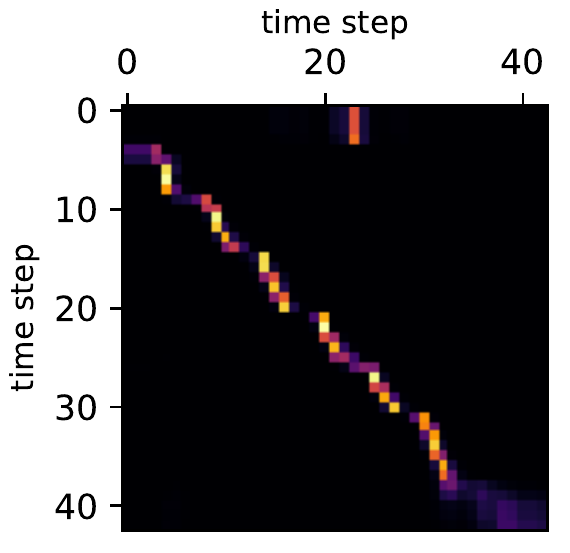}
	}	
	\quad
	\subfloat[The attention scores of the third head of the last encoder layer.]
	{   \label{fig:enc_self_att_layer5_head2}
		\includegraphics[width=0.21\linewidth]{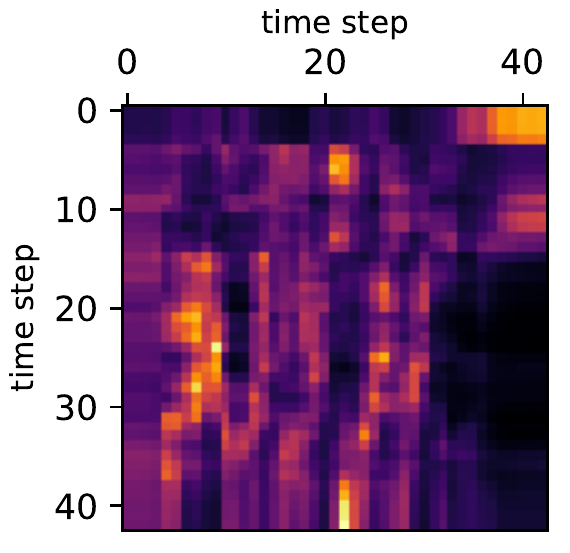}
	}	
	\quad
	\subfloat[The attention scores of the forth head of the last encoder layer.]
	{   \label{fig:enc_self_att_layer5_head3}
		\includegraphics[width=0.21\linewidth]{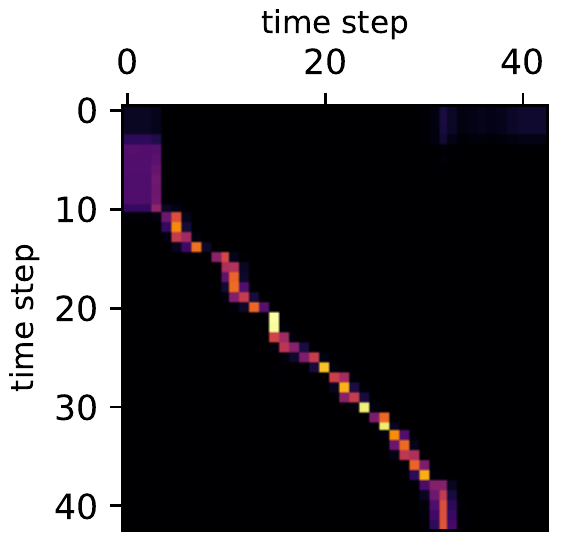}
	}	
	
	\subfloat[The attention scores of the first head of the last decoder layer.] 
	{   \label{fig:dec_att_layer5_head0}
		\includegraphics[width=0.21\linewidth]{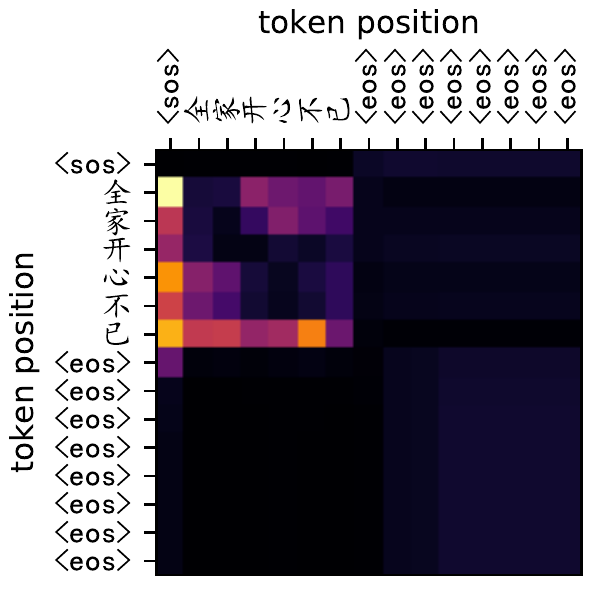}
	}
	\quad	
	\subfloat[The attention scores of the second head of the last decoder layer.] 
	{   \label{fig:dec_att_layer5_head1}
		\includegraphics[width=0.21\linewidth]{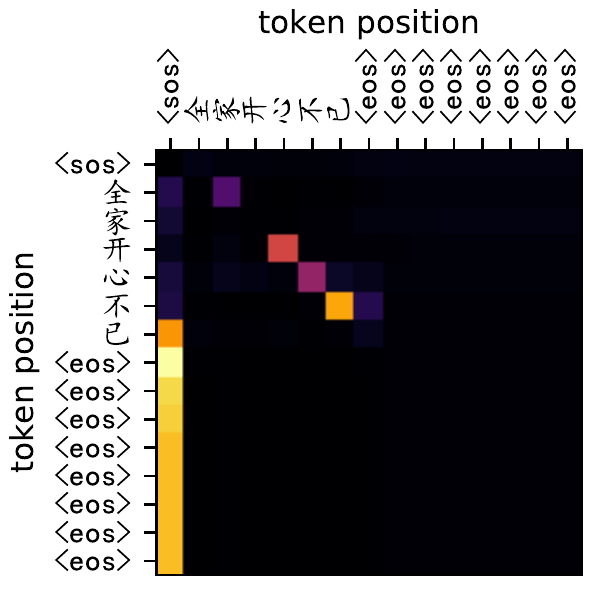}
	}	
	\quad
	\subfloat[The attention scores of the third head of the last decoder layer.] 
	{   \label{fig:dec_att_layer5_head2}
		\includegraphics[width=0.21\linewidth]{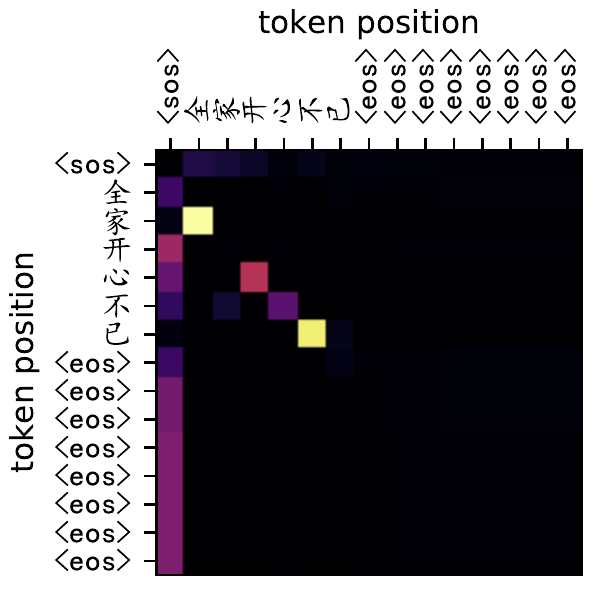}
	}	
	\quad
	\subfloat[The attention scores of the forth head of the last decoder layer.] 
	{   \label{fig:dec_att_layer5_head3}
		\includegraphics[width=0.21\linewidth]{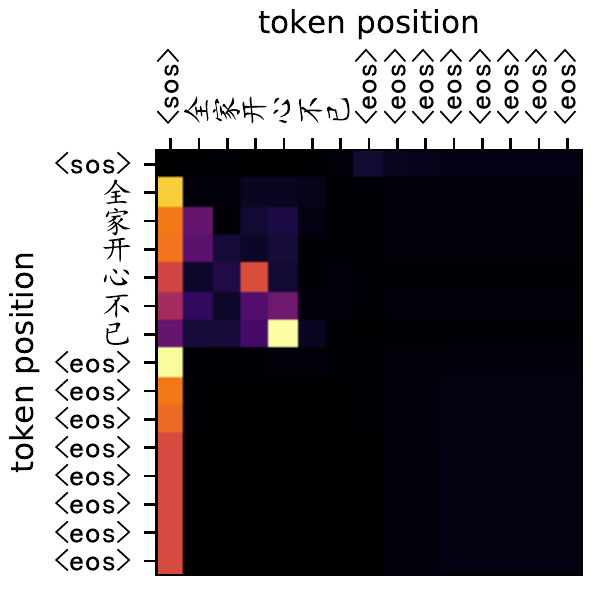}
	}	
	
	\subfloat[The attention scores of the first head of the last PDS layer.] 
	{   \label{fig:pds_att_layer1_head0}
		\includegraphics[width=0.21\linewidth]{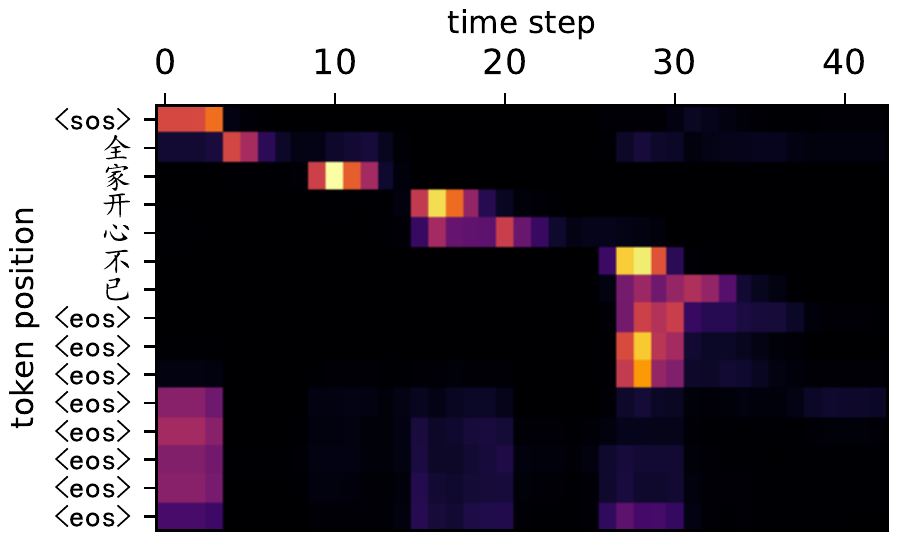}
	}
	\quad	
	\subfloat[The attention scores of the second head of the last PDS layer.] 
	{   \label{fig:pds_att_layer1_head1}
		\includegraphics[width=0.21\linewidth]{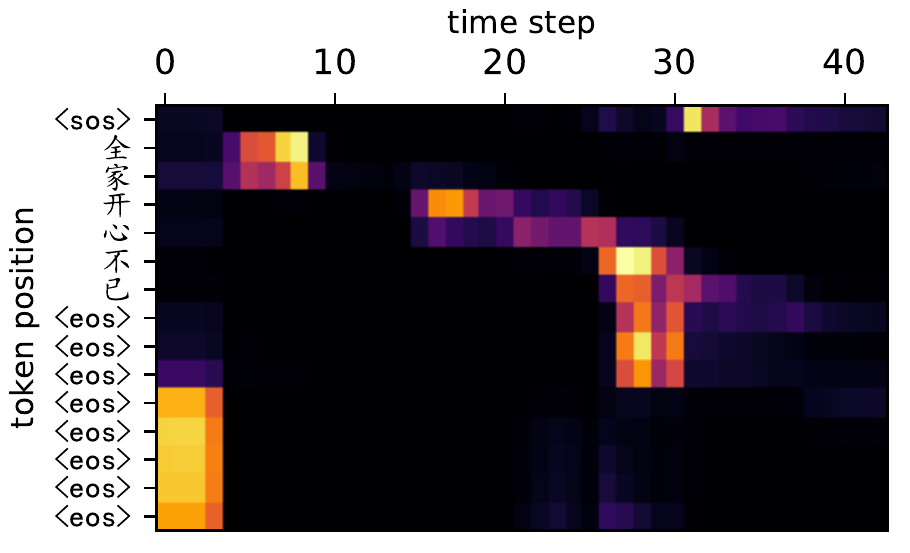}
	}
	\quad	
	\subfloat[The attention scores of the third head of the last PDS layer.] 
	{   \label{fig:pds_att_layer1_head2}
		\includegraphics[width=0.21\linewidth]{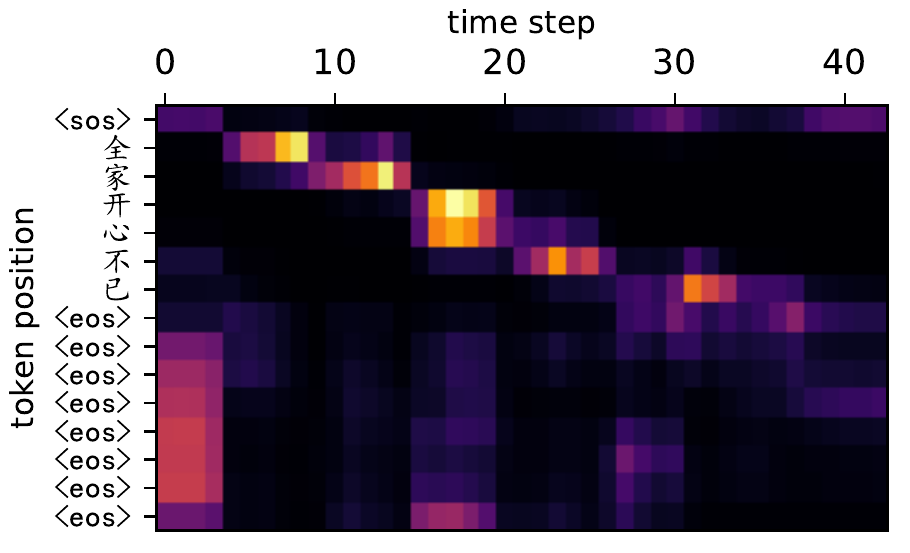}
	}	
	\quad
	\subfloat[The attention scores of the forth head of the last PDS layer.] 
	{   \label{fig:pds_att_layer1_head3}
		\includegraphics[width=0.21\linewidth]{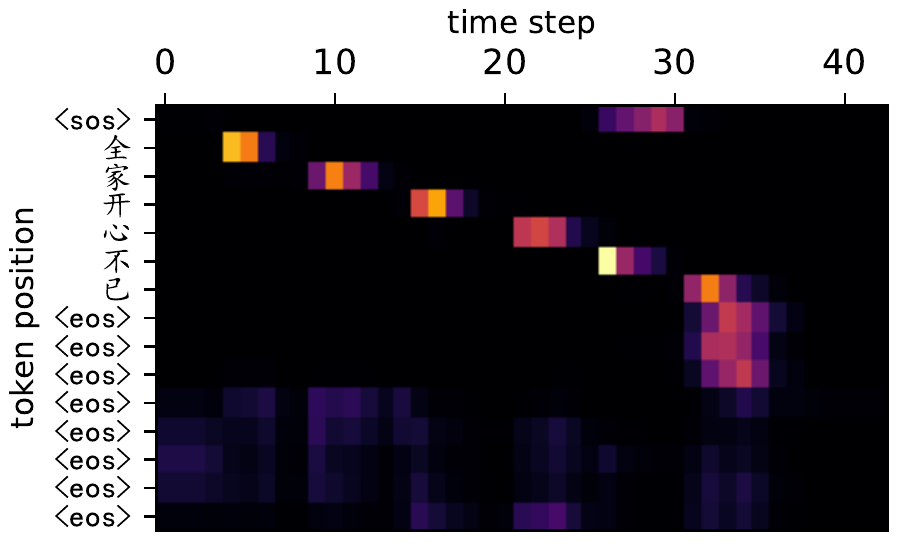}
	}	
	\caption{The visualization of the attention scores of the model \texttt{LASO-big}. Here, we show the first four heads of the last layer of each module. And because the token sequence is long to visualize, we truncate the first 15 tokens. All visualization is listed in the supplemental materials.}
	\label{fig:vis} 
	\vspace{-10pt}
\end{figure*}

\vspace{-10pt}
\subsection{Comparisons with Other Methods on AISHELL-2}

We then extend the experiments to the larger scale dataset AISHELL-2. AISHELL-2 contains about 1000 hours of training data. And the covered topics of AISHELL-2 are more diverse than AISHELL-1. Furthermore, the training set is recorded with iPhone, and the test sets of AISHELL-2 cover three different channels, i.e., iPhone, Android smartphones, and Hi-Fi microphones. So we can evaluate the generalization of the models on AISHELL-2 more in detail. Because AISHELL-2 is larger than AISHELL-1 and training models on AISHELL-2 costs much more time, we directly use the selected architectures from previous experiments on this dataset.

The experimental results are shown in \autoref{tab:compare_aishell2}. We can see that the proposed LASO models also achieve a promising performance. Compared with the hybrid systems, LAS systems, and the CTC model, all LASO models achieve better performance. And with the similar and larger scale model sizes, \texttt{LASO-middle} and \texttt{LASO-big} outperform previous transformer models. We find that with more data and a larger model, LASO can achieve better performance than the well-trained transformer models. With semantic refinement from BERT, the performances are further improved. However, the improvements are not significant like the experiments on AISHELL-1. And for \texttt{LASO-small}, the performance degrades. The possible reasons are 1) the model capacity is very large between the small model \texttt{LASO-small} and the big model BERT; 2) AISHELL-2 has more data and is more complex than AISHELL-1, which influences the effectiveness of knowledge distillation \cite{Zagoruyko2017AT,cho2019efficacy}.

\vspace{-5pt}
\section{Discussion}
\label{sec:discuss}
In this section, we analyze the attention patterns and discuss the impact of the sentence lengths. And we show some decoded examples of the models.

\begin{figure}[!t]\centering
	\subfloat[A scatter plot of the CER of each sentence vs. the length of a sentence on AISHELL-1 test set.] 
	{
		\includegraphics[width=0.85\linewidth]{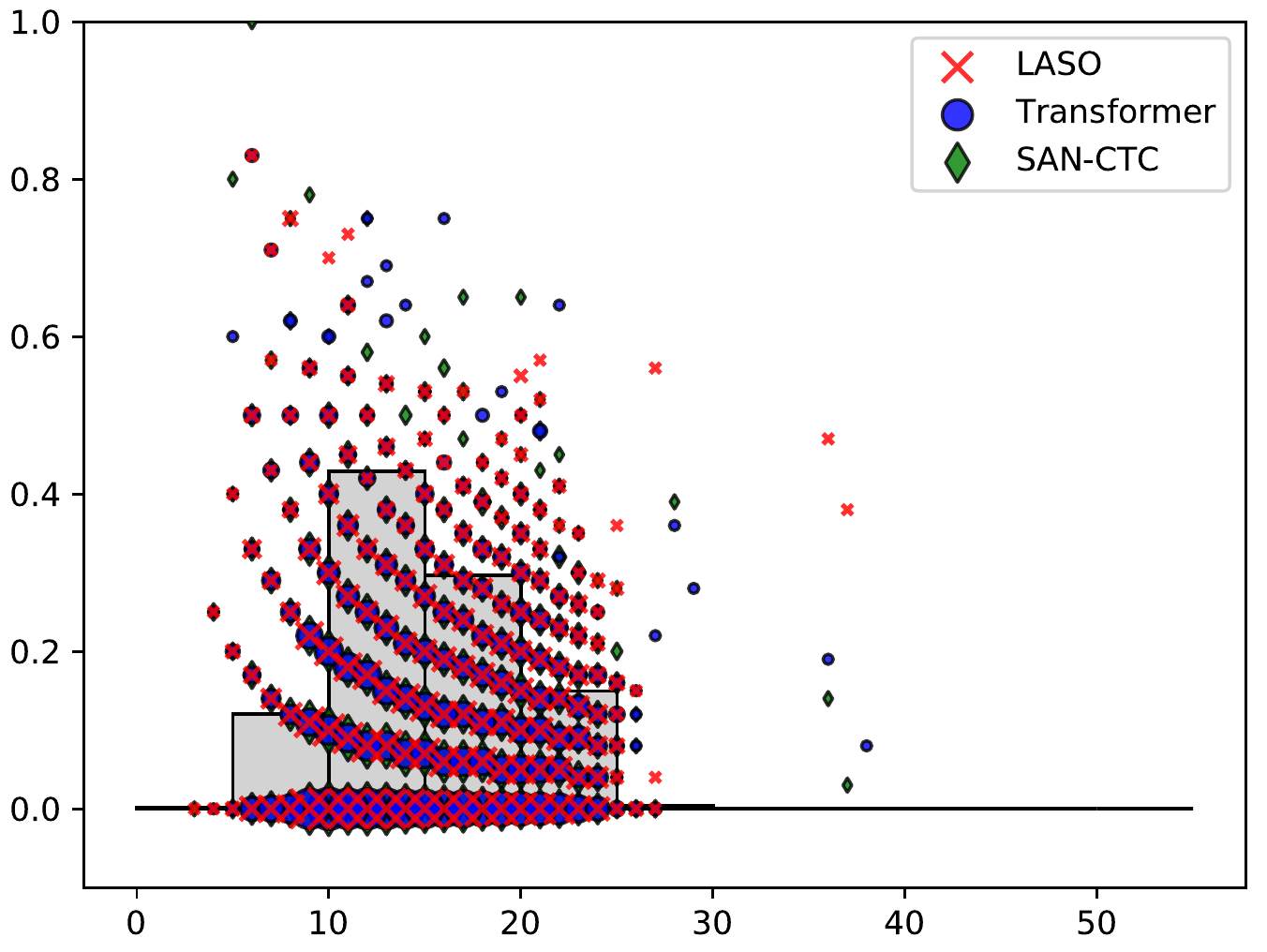}
	}	
	
	\subfloat[A scatter plot of the CER of each sentence vs. the length of a sentence on AISHELL-2 iPhone test set.] 
	{   
		\includegraphics[width=0.85\linewidth]{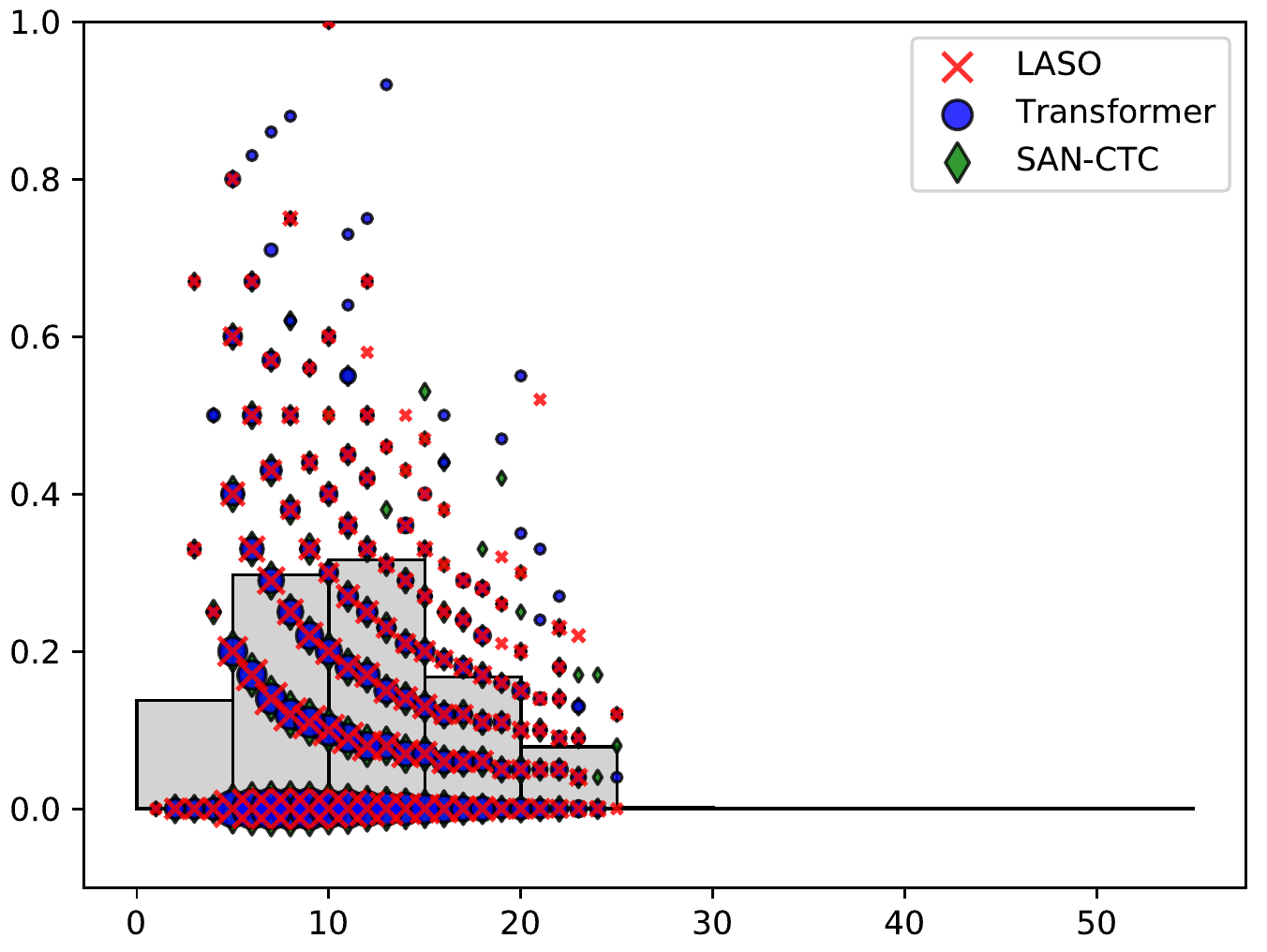}
	}
	\caption{Scatter plots for analyzing the impact of the lengths of the sentences. The grey histograms represent the ratios of lengths of the sentence (i.e., the number of the tokens) on the training set. The area of a scatter shows the number of the examples at that point. We use baseline1 \texttt{Transformer}, baseline2 \texttt{SAN-CTC}, and \texttt{LASO-middle} since they have similar model sizes. From the plots, we can see no significant difference between the three models. The three models have the same CER for some sentences so that the scatters are overlapped. Note that an outlier in the second figure which CER is 100\% since the reference of this sentence is wrong. }
	\label{fig:length}
	\vspace{-10pt}
\end{figure}

\subsection{Visualization of Attention Patterns}

To better understand the behaviors of the LASO model, we visualize the attention scores of an utterance with \texttt{LASO-big}. We show the first four heads of attention scores of the last layers of the encoder, the PDS, and the decoder. The detailed visualization is listed in the supplemental materials. \autoref{fig:vis} shows the visualization results. We summarize the observations as follows.

\begin{enumerate}
	\item Different heads in one layer have different attention patterns. This implies that one representation of the sequence in different head attends different representations.
	\item For the encoder, some attention patterns show the top-left to bottom-right alignments (\autoref{fig:enc_self_att_layer5_head1} and \autoref{fig:enc_self_att_layer5_head3}). This meets the expectation: the matched representations are around the corresponding representation in the speech sequence. And some heads do not show obvious patterns.
	\item For the decoder, we can see that some hidden representation at a token position attends the hidden representation at the previous token position (\autoref{fig:dec_att_layer5_head2}), and some hidden representation at a token position attends the hidden representation at the next token position (\autoref{fig:dec_att_layer5_head1}). Most hidden representations corresponding to filler tokens \texttt{<eos>} attend the hidden representations corresponding to \texttt{<sos>}.
	\item For the PDS, the position corresponding to a token attends a small range of representations, and the overall patterns are from the top-left to the bottom-right (\autoref{fig:pds_att_layer1_head0}, \autoref{fig:pds_att_layer1_head1}, \autoref{fig:pds_att_layer1_head2}, and \autoref{fig:pds_att_layer1_head3}). For the first several positions corresponding to the filler \texttt{<eos>} tokens, the attention pattern is like a vertical line. However, the lines are corresponding to different positions of the speech in different head (\autoref{fig:pds_att_layer1_head1} and \autoref{fig:pds_att_layer1_head3}). And most other \texttt{<eos>} tokens attend the first several representations. We analyze that these representations denote fillers.	
\end{enumerate}

From the above observations, we conclude that 1) different attention patterns make the model fuse the representations from various aspects; 2) the attention mechanism can learn meaningful alignments in terms of the positional encodings; 3) the two special filler tokens \texttt{<sos>} and \texttt{<eos>} absorb the meaningless representations from the encoder; 4) specific attention patterns exist in the different heads of the decoder. These demonstrate that the PDS module attends specific acoustic representations based on positions and the decoder captures the token relationship based on the self-attention mechanism.

\vspace{-10pt}
\subsection{The Impact of the Lengths of Sentences}
Position parameters exist in the PDS module. This may cause that the performance relies on the length of a sentence. To check this point, we plot scatters of the CER of each sentence vs. the length of a sentence in \autoref{fig:length}. We can see \textit{no significant} difference among the three models.

\autoref{fig:length} also provides histograms of ratios of the different lengths of the training set. We can see that the distributions approximate Gaussian distribution as expected. And the trend of CERs is similar to the length distribution of the training set, i.e., the CERs of more sentences are zero in the middle part of the figures. This is because the models are trained with more data with the middle lengths. All three models can recognize the sentences with unseen lengths (or few-shot lengths) in the training set. However, the error rates are relatively higher than the sentences with seen lengths. This phenomenon is as expected. Because the three models (\texttt{Transformer}, \texttt{SAN-CTC}, \texttt{LASO}) are all whole-utterance ASR models, the length of the utterance is an implicit factor to train the models. This is different from the local-window models based conventional hybrid models, i.e., time-delay neural networks (TDNN), CNNs, or latency-control BLSTM. LASO is more sensitive with the unseen lengths than \texttt{Transformer} and \texttt{SAN-CTC}. Because the PDS module needs to be trained with various lengths. 

In engineering practice, two methods can be applied to address the unseen-length issue of the whole-utterance models: 1) select data with various lengths to train the models; 2) using a voice activity detection (VAD) system to cut long utterances into segments.        

\vspace{-10pt}
\subsection{Examples of Semantic Refinement from BERT}
We show some recognized results of the models with and without semantic refinement from BERT in \autoref{tab:example}. The recognized results are from AISHELL-1 test set. The model architecture is \texttt{LASO-big}. The grey background is the wrongly recognized words. And the red words are correctly recognized by the model with semantic refinement from BERT.

From \autoref{tab:example}, we can see that in terms of the semantic of the words, the model with semantic refinement from BERT correctly recognized the words which were not recognized by the model without semantic refinement from BERT. Specifically, for case 1 to case 4, the model w/o BERT recognized the word as another word which has similar pronunciation but mismatched meaning (e.g., ``\begin{CJK}{UTF8}{gkai}一场\end{CJK}'', ``\begin{CJK}{UTF8}{gkai}骤\end{CJK}''). Case 5 is an interesting example. The model w/o BERT recognized a word as ``\begin{CJK}{UTF8}{gkai}爷泳\end{CJK}'', but it is wrong in grammar. The results of the model w/ BERT at the corresponding position is ``\begin{CJK}{UTF8}{gkai}爷爷\end{CJK}'' which is a correct word in grammar, but it does not match the reference. These results show that the proposed semantic refinement from BERT impacts the model in semantic and can improve semantic representation performance for these examples.

\section{Conclusions and Future Work}
\label{sec:conc}
This paper proposes a feedforward neural network based non-autoregressive speech recognition model called LASO. The model consists of an encoder, a position dependent summarizer, and a decoder. The encoder encodes the acoustic feature sequence into high-level representations. The PDS converts the acoustic representation sequence to the token-level sequence. And the decoder further captures the token-level relationship. Because the prediction of each token does not rely on other tokens, and the whole model is feedforward, the parallelization of the whole sentence prediction is realizable. Thus, the inference speed is much improved, compared with the autoregressive attention-based end-to-end models. Furthermore, we propose to refine semantics from a large-scale pre-trained language model BERT to improve the performance. Experimental results show that LASO can achieve a competitive performance and high efficiency. In the future, we will improve the performance of LASO by the architecture and loss functions. We find that LASO is trained with more epochs during training. We will try to find strategies to speed up training. And we will try to find more proper modeling units for Latin-alphabet-based data.

\begin{table}[!t]	
\setlength\tabcolsep{4pt}
	\caption{Examples of Recognized Results}
	\centering
	\begin{CJK}{UTF8}{gkai}
	\begin{threeparttable}
	\begin{tabular}{l|l|l}	
	\hline  
	\multirow{3}{*}{1} & Reference& 而二零零八年举办夏季奥运会所留下的宝贵\stackon{遗产}{/yi2 chan3/} \\ \cline{2-3}
					   \\[-1em] 
					   & w/o BERT & 而二零零八年举办夏季奥运会所留下的宝贵\hl{\stackon{一场}{/yi4 chang3/}} \\ \cline{2-3}
					   \\[-1em]
					   & w/ BERT  & 而二零零八年举办夏季奥运会所留下的宝贵\textcolor{red}{\stackon{遗产}{/yi2 chan3/}} \\ \hline
					   \\[-1em]
	\multirow{3}{*}{2} & Reference& 当月住宅类商品房成交套数\stackon{骤}{/zhou4/}跌 \\ \cline{2-3}
					   \\[-1em] 
				       & w/o BERT & 当月住宅类商品房成交套数\hl{\stackon{周}{/zhou1/}}跌 \\ \cline{2-3}
				       \\[-1em] 
					   & w/ BERT  & 当月住宅类商品房成交套数\textcolor{red}{\stackon{骤}{/zhou4/}}跌 \\ \hline
					   \\[-1em]	
	\multirow{3}{*}{3} & Reference& 数十名市民赶到越秀区\stackon{一}{/yi4/}酒\stackon{家维权}{/jia1 wei2 quan2/} \\ \cline{2-3}
					   \\[-1em] 
					   & w/o BERT & 数十名市民赶到越秀区\hl{\stackon{以}{/yi3/}}酒\hl{\stackon{酒未钱}{/jiu3 wei4 qian2/}} \\ \cline{2-3}
					   \\[-1em] 
					   & w/ BERT  & 数十名市民赶到越秀区\textcolor{red}{\stackon{一}{/yi4/}}酒\textcolor{red}{\stackon{家维权}{/jia1 wei2 quan2/}} \\ \hline
					   \\[-1em]		
	\multirow{3}{*}{4} & Reference& 尽管她努力\stackon{瘦身}{/shou4 shen1/} \\ \cline{2-3} 
					   \\[-1em]
					   & w/o BERT & 尽管她努力\hl{\stackon{受存}{/shou4 cun2/}} \\ \cline{2-3}
					   \\[-1em] 
					   & w/ BERT  & 尽管她努力\textcolor{red}{\stackon{瘦身}{/shou4 shen1/}} \\ \hline
					   \\[-1em]
	\multirow{3}{*}{5} & Reference& \stackon{圆圆}{/yuan2 yuan2/}的脸蛋非常的可爱	 \\ \cline{2-3}
					   \\[-1em] 
					   & w/o BERT & \hl{\stackon{爷泳}{/ye2 yong3/}}的脸蛋非常的可爱 \\ \cline{2-3}
					   \\[-1em] 
					   & w/ BERT  & \hl{\stackon{爷爷}{/ye2 ye5/}}的脸蛋非常的可爱 \\ \hline					   		
	\end{tabular}
	\footnotesize
	$/\cdot/$ is pinyin for labeling the pronunciation of the Chinese characters. The number is the tone of the character. 
	\end{threeparttable}
	\end{CJK}
\label{tab:example}
\vspace{-10pt}
\end{table}	

\section{Acknowledgment}
The authors are grateful to the anonymous reviewers for their invaluable comments that improve the completeness and readability of this paper.

\ifCLASSOPTIONcaptionsoff
  \newpage
\fi



%




\bibliographystyle{IEEEtran}
\bibliography{refs}{}

	

%




\end{document}